\documentclass[pdflatex,sn-nature]{sn-jnl}

\usepackage{graphicx}%
\usepackage{multirow}%
\usepackage{amsmath,amssymb,amsfonts}%
\usepackage{amsthm}%
\usepackage{mathrsfs}%
\usepackage[title]{appendix}%
\usepackage{xcolor}%
\usepackage{textcomp}%
\usepackage{manyfoot}%
\usepackage{booktabs}%
\usepackage{algorithm}%
\usepackage{algorithmicx}%
\usepackage{algpseudocode}%
\usepackage{listings}%
\usepackage{lineno}
\usepackage{multirow}
\usepackage{subcaption}%
\usepackage{placeins}
\usepackage{booktabs}
\usepackage{float}
\usepackage{svg}
\usepackage{booktabs}
\usepackage{multirow}
\usepackage{graphicx}
\usepackage{longtable} % Include this package in the preamble
\usepackage{lineno}
\usepackage{setspace}
\theoremstyle{thmstyleone}%
\theoremstyle{thmstyletwo}%

\theoremstyle{thmstylethree}%

\begin{document}

\title[Unraveling Groundwater Stress]{Unraveling Groundwater Stress: A Model Generalizable Across Space and Time Reveals Co-Declining Water Tables and Recharge Across Bangladesh}
\author*[1]{\fnm{Saleh Sakib} \sur{Ahmed}}
\email{salehsakibahmed@gmail.com}

\author[2]{\fnm{Rashed Uz} \sur{Zzaman}}
\email{shoourov011@gmail.com}

\author[1]{\fnm{Sheikh Saifur Rahman} \sur{Jony}}
\email{srj.buet17@gmail.com}

\author[2]{\fnm{Faizur Rahman} \sur{Himel}}
\email{himel.faizur@gmail.com}

\author[3]{\fnm{Afroza} \sur{Sharmin}}
\email{asharmin32@yahoo.com}

\author[4]{\fnm{A.H.M. Khalequr} \sur{Rahman}}
\email{hasnat.dphe@gmail.com}

\author*[1]{\fnm{M. Sohel} \sur{Rahman}}
\email{sohel.kcl@gmail.com}

\author*[2]{\fnm{Sara} \sur{Nowreen}}
\email{snowreen@iwfm.buet.ac.bd}

\affil[1]{\orgdiv{Department of Computer Science and Engineering},
\orgname{Bangladesh University of Engineering and Technology (BUET)},
\city{Dhaka}, \country{Bangladesh}}

\affil[2]{\orgname{Institute of Water and Flood Management},
\orgname{Bangladesh University of Engineering and Technology (BUET)},
\city{Dhaka}, \country{Bangladesh}}

\affil[3]{\orgname{Bangladesh Agricultural Development Corporation (BADC)},
\city{Dhaka}, \country{Bangladesh}}

\affil[4]{\orgname{Department of Public Health Engineering (DPHE)},
\city{Dhaka}, \country{Bangladesh}}

\abstract{
Is Bangladesh replenishing the groundwater it depends on? Answering this question remains difficult in Bangladesh and worldwide because sparse \textit{in-situ} measurements limit observational capabilities, while existing machine-learning approaches struggle to generalize across time and space. To address this bottleneck, we develop a Teacher--Student framework in which a Teacher accurately learns from sparse annual \textit{in-situ} minimum (\(R^2 = 0.86\)) and maximum (\(R^2 = 0.96\)) depth observations to generate dense spatial patterns, and a Student transfers these patterns from time-varying 25-km GLDAS groundwater storage to high-resolution (2-km) depth maps using location-specific features. The Student achieves \(R^2 = 0.93\) over 2003--2022 and generalizes across unseen years (Leave-One-Year-Out \(R^2 = 0.93\)) and locations (Leave-One-District-Out \(R^2 = 0.70\)), enabling reconstruction without contemporary \textit{in-situ} measurements. The reconstruction reveals widespread water-table deepening and a national decline in groundwater recharge of \(1.10~\text{mm yr}^{-1}\) (\(p < 0.001\)). Autocorrelation-corrected analysis identifies statistically significant recharge declines across 55.6\% of Bangladesh, compared with significant increases across only 4.9\%. This imbalance suggests that pumping-induced recharge enhancement, as proposed by the Bengal Water Machine hypothesis, remains localized rather than widespread across Bangladesh. Together, these findings reveal widespread weakening of groundwater replenishment and provide a scalable framework for monitoring groundwater dynamics in data-sparse aquifers globally.
}




\maketitle

\section{Introduction}\label{sec1}

Assessing whether aquifers are replenished fast enough to sustain human extraction remains a fundamental challenge, particularly in data-sparse regions. Driven by seasonal surface-water scarcity, South Asia's reliance on shallow aquifers has surged, with tube wells in Bangladesh multiplying 15-fold since 1981 \cite{qureshi2014groundwater,harvey2006groundwater}. In the face of this intense extraction, the sustainability of these aquifers hinges on groundwater recharge, which can be quantified from annual water-table fluctuations and soil \textit{specific yield} via the Water Table Fluctuation (WTF) method \cite{bonsor2017hydrogeological,delottier2018estimating,addisie2022groundwater}. Consequently, this makes multi-year groundwater-level maps essential for assessing recharge trends and emerging depletion \cite{tleane2020estimation,tadesse2024assessing}. Yet \textit{in-situ} well networks, while accurate, are spatially sparse and costly to maintain, making multi-year, national-scale groundwater reconstruction difficult. Here, we address this challenge by introducing a generalizable Teacher--Student framework that resolves multi-year groundwater dynamics across space and time using satellite-derived groundwater storage data without requiring contemporary monitoring wells.

Existing groundwater modelling approaches often struggle to generalize across both space and time. Sparse observations are commonly interpolated using methods such as Inverse Distance Weighting (IDW)~\cite{shepard1968two}, Kriging~\cite{wackernagel2003ordinary}, Empirical Bayesian Kriging (EBK)~\cite{krivoruchko2012empirical}, and M5P~\cite{wang1997inducing}, while some studies use these estimates as inputs to machine-learning models \cite{verma2020groundwater,abdi2024enhancing}. However, interpolation can smooth localized hydrogeological variability and become unreliable as monitoring networks become sparse, potentially propagating spatial biases into downstream models \cite{mitas1999spatial,huang2024ground,esri2012ebk}. Machine-learning models capture nonlinear relationships, but their generalizability depends on predictor choice. Models relying on static attributes (e.g., elevation and soil type)~\cite{zzaman2023cimca,zhang2021downscaling} lack temporal variation, whereas slowly changing remotely sensed surface indices—such as the Normalized Difference Vegetation Index (NDVI) and Normalized Difference Water Index (NDWI)—provide limited information on interannual groundwater variability~\cite{ukasha2022temporal,hu2023spatiotemporal,ashok2021monitoring,nagkoulis2025modeling}. Rainfall and evapotranspiration capture temporal variation, but their effects on groundwater can be delayed and vary spatially with geographical properties (e.g. soil, lithology, and land use) \cite{santos2017space,feng2024predicting,zeng2022groundwater,panda2024exploring,igwebuike2025application}. Consequently, many models are either trained for specific years~\cite{zhang2021downscaling,zzaman2023cimca} or formulated as time-series models~\cite{zeng2022groundwater,panda2024exploring,igwebuike2025application} using these features, limiting their ability to transfer to unseen years or locations. Moreover, most studies focus on daily or monthly prediction or on localized applications, leaving multi-year, annual-scale groundwater reconstruction comparatively underexplored \cite{tao2022groundwater,khan2023comprehensive}. Thus, a framework that produces spatially continuous annual groundwater maps while generalizing across both unseen years and locations remains a critical missing link.

To find such a method, rather than looking downwards, this study looks to a different source above us. Orbiting Earth, the twin satellites of the \textit{Gravity Recovery and Climate Experiment (GRACE)} mission measure changes in Earth's gravity field, from which changes in total terrestrial water storage are estimated.  These observations are assimilated into the \textit{Global Land Data Assimilation System (GLDAS-2.2)}, a land-surface modelling framework, to produce model-based estimates of groundwater-storage variability at approximately (25~\text{km}) resolution. Although these estimates capture temporal changes in regional groundwater storage, their coarse resolution cannot resolve local groundwater-depth variations over distances of only a few kilometres \cite{brammer2012physical}. The challenge, therefore, is to translate this large-scale, time-varying storage information into fine-scale groundwater-depth maps.

% Specifically, our framework bridges this resolution gap by downscaling coarse (\(25~\text{km}\)) GLDAS groundwater storage into high-resolution (\(2~\text{km}\)) annual groundwater-depth maps across Bangladesh. The Teacher learns depth patterns from sparse \textit{in-situ} observations and hydrogeological characteristics, generating dense \(2~\text{km}\) maps to supervise a generalizable Student model. Guided by this dense supervision, the Student learns the underlying dynamics: given the groundwater storage and environmental conditions of a broader \(25~\text{km}\) region, it determines the exact depth at each \(2~\text{km}\) location based on local site features. This allows the Student to use coarse, time-varying storage data to accurately estimate depths at locations where no \textit{in-situ} measurements exist. Applied across Bangladesh over 2003--2022 and extrapolated to 2023--2024, the framework achieves an overall \(R^2=0.93\), with strong temporal (Leave-One-Year-Out \(R^2=0.93\)) and spatial (Leave-One-District-Out \(R^2=0.70\)) generalization.

Specifically, our framework bridges this resolution gap by downscaling coarse (\(25~\text{km}\)) GLDAS groundwater storage into high-resolution (\(2~\text{km}\)) annual groundwater-depth maps across Bangladesh. The Teacher accurately captures depth patterns from sparse \textit{in-situ} observations and hydrogeological characteristics, generating dense \(2~\text{km}\) maps to supervise the generalizable Student model. Using this dense supervision, the Student dynamically links regional scales to local scales: given the groundwater storage and environmental conditions of a broader \(25~\text{km}\) region, it determines the exact depth at a specific \(2~\text{km}\) location based on its local site features. Consequently, the framework can continuously reconstruct time-varying groundwater depths even in locations entirely lacking \textit{in-situ} measurements. Applied across Bangladesh over 2003--2022 and extrapolated through 2024, the model achieves an overall \(R^2=0.93\) alongside robust temporal (Leave-One-Year-Out \(R^2=0.93\)) and spatial (Leave-One-District-Out \(R^2=0.70\)) generalization.

The resulting groundwater-depth and recharge maps reveal widespread long-term water-table deepening, with pronounced declines in northwestern and central Bangladesh. Recharge is declining significantly across 55.6\% of the country, while significant positive trends occur in only 4.9\%. These results provide a spatially resolved view of the Bengal Water Machine hypothesis~\cite{shamsudduha2022bengal}, which proposes that intensive dry-season pumping lowers water tables, creating subsurface storage that captures additional monsoon rainfall and enhances recharge. Our results demonstrate that this recharge enhancement is highly localized rather than a nationwide phenomenon, likely restricted to areas with favorable hydrogeological conditions. Ultimately, by resolving groundwater dynamics across space and time, our framework provides a critical basis for assessing groundwater sustainability across Bangladesh and offers a scalable approach for tracking groundwater dynamics in other data-sparse aquifers globally.

\begin{figure}
    \centering
    \includegraphics[width=\textwidth]{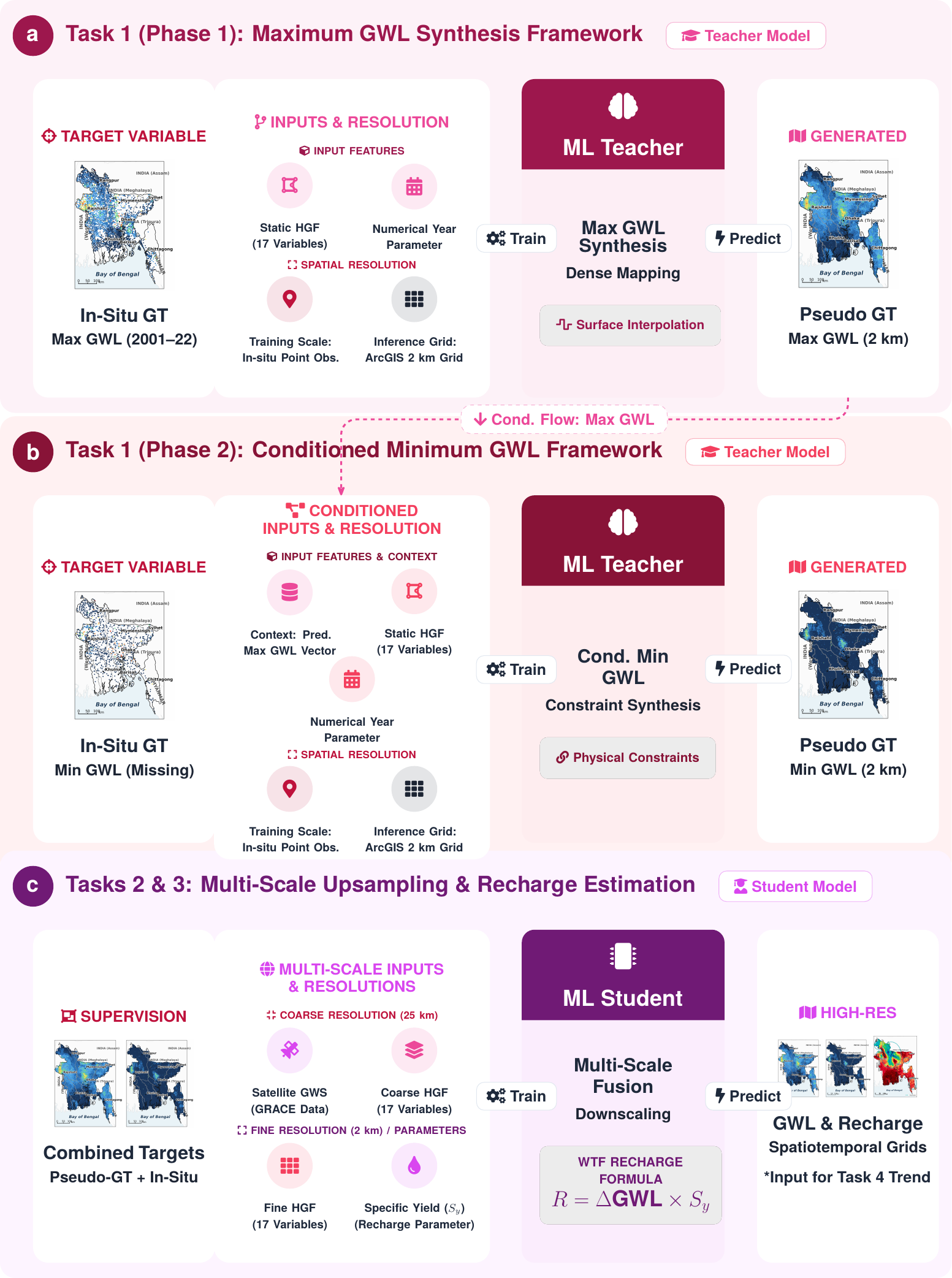}
    \caption{\textbf{Methodology:} \textbf{a.} Phase 1: Teacher Model predicting maximum groundwater level (bgl) using 17 HGFs and year as inputs. \textbf{b.} Phase 2: Teacher Model predicting minimum groundwater level (bgl) conditioned on maximum groundwater level (bgl) to ensure consistency, with inputs including 17 HGFs, year, and maximum groundwater level (bgl). \textbf{c.} Upsampling Student Model combining grid-level low-resolution information (annual maximum and minimum groundwater storage along with 17 HGFs of grid center-point) and point-level high-resolution data (17 HGFs of 2~km), merged by GLDAS grid ID and year, to predict high-resolution annual maximum and minimum groundwater level (bgl). The target for the Student is the Teacher's pseudo-labels and the original in-situ data}

    \label{fig:method}
\end{figure}

\section{Results}
\subsection{Overview of Methods}

Our framework executes four primary tasks within a \textit{Teacher--Student} paradigm (Fig.~\ref{fig:method}). Task 1 generates the dense spatial supervision required to train the Student model through a two-phase process. Phase 1 pools \textit{in-situ} observations from 2001--2022 to train a Teacher model that reconstructs annual maximum groundwater depth. This model leverages 17 hydrogeological features (HGFs; detailed in the Methods) alongside a numerical year encoding to capture temporal variability. Phase 2 reconstructs minimum groundwater depth using the same inputs as Phase 1, while explicitly conditioned on Phase 1 outputs to physically constrain the minimum depth from exceeding the maximum. Once trained, both Teacher models generate dense annual groundwater-depth maps across a \(2~\text{km}\) grid. These dense predictions serve as pseudo-ground truth; combined with the actual \textit{in-situ} ground truth, they provide the complete spatial supervision required to train the generalizable Student model.

Task 2 trains the Student model (GroundHog) to dynamically reconstruct groundwater depth at \(2~\text{km}\) resolution from time-varying annual \(25~\text{km}\) GLDAS groundwater-storage (GWS) estimates, alongside regional (\(25~\text{km}\)) and local (\(2~\text{km}\)) HGFs. Task 3 applies the Water Table Fluctuation method to these reconstructed minimum and maximum depths, integrating specific yield to estimate annual groundwater recharge. Finally, Task 4 quantifies long-term spatiotemporal trends in groundwater depth and recharge (2003--2022) using Sen's slope estimator \cite{sen1968estimates} alongside standard \cite{MannKendallTest} and Modified \cite{hamed1998modified, yue2002modified} Mann--Kendall tests, with the latter explicitly correcting for temporal autocorrelation to prevent exaggerated statistical significance.

\begin{table}[htbp]
\centering
\small
\caption{Comprehensive Performance Comparison of Teacher Model Candidates for Task 1 (Max and Min GWL Prediction). \textbf{Section A} evaluates model performance under year-specific training constraints. \textbf{Section B} compares combined-year training architectures, identifying Random Forest as the top-performing ML backbone. \textbf{Section C} benchmarks the proposed Groundhog framework (Combined-Year Random Forest) against traditional spatial interpolation baselines.}
\label{tab:teacher_candidates_task1}
\begin{tabular}{llccccc}
\toprule
\textbf{Category / Experiment} & \textbf{Model / Framework} & $\mathbf{R^2}$ & \textbf{MSE} & \textbf{RMSE} & \textbf{MAE} & \textbf{MAPE (\%)} \\
\midrule
\multicolumn{7}{l}{\textbf{A. Year-Specific Architecture Comparison (Ablation Study)}} \\
\midrule
\multirow{4}{*}{Max GWL} 
 & Random Forest        & \textbf{0.3115} & \textbf{22.8765} & \textbf{4.7068} & \textbf{2.3979} & \textbf{34.20} \\
 & Neural Network       & 0.1993          & 26.5521          & 5.0912          & 2.7973          & 41.13 \\
 & XGBoost              & 0.2172          & 25.9802          & 5.0143          & 2.6763          & 39.29 \\
 & LightGBM             & 0.2688          & 24.3531          & 4.8523          & 2.6511          & 38.78 \\
\cmidrule(lr){2-7}
\multirow{4}{*}{Min GWL} 
 & Random Forest        & \textbf{0.1082} & \textbf{33.1914} & \textbf{5.6804} & \textbf{3.0257} & 82.76 \\
 & Neural Network       & 0.0771          & 36.0411          & 5.8703          & 3.1026          & \textbf{75.54} \\
 & XGBoost              & -0.0685         & 39.6183          & 6.1848          & 3.2957          & 95.97 \\
 & LightGBM             & 0.0322          & 34.7200          & 5.8341          & 3.3360          & 97.62 \\

\midrule
\multicolumn{7}{l}{\textbf{B. Combined-Year Architecture Comparison (Model Selection)}} \\
\midrule
\multirow{4}{*}{Max GWL} 
 & Random Forest        & \textbf{0.8767} & \textbf{3.4468} & \textbf{1.6590} & \textbf{0.8824} & \textbf{13.95} \\
 & Neural Network       & 0.2024          & 23.5689         & 4.7940          & 2.7625          & 47.05 \\
 & XGBoost              & 0.8133          & 5.2908          & 2.1905          & 1.4010          & 23.66 \\
 & LightGBM             & 0.6924          & 8.9252          & 2.9348          & 1.8514          & 32.70 \\
\cmidrule(lr){2-7}
\multirow{4}{*}{Min GWL} 
 & Random Forest        & \textbf{0.9668} & \textbf{1.1866} & \textbf{1.0480} & \textbf{0.6958} & 81.57 \\
 & Neural Network       & 0.7831          & 7.4989          & 2.7079          & 2.0069          & 266.36 \\
 & XGBoost              & 0.9636          & 1.3823          & 1.1118          & 0.7355          & \textbf{72.89} \\
 & LightGBM             & 0.9518          & 1.8142          & 1.2952          & 0.8786          & 105.65 \\

\midrule
\multicolumn{7}{l}{\textbf{C. Proposed Framework vs. Spatial Interpolation Baselines}} \\
\midrule
\multirow{4}{*}{Max GWL} 
 & Groundhog (Proposed) & \textbf{0.8542} & \textbf{4.1913} & \textbf{2.0473} & \textbf{0.9712} & \textbf{14.54} \\
 & IDW                  & 0.1423          & 44.9605         & 6.5928          & 3.2319          & 41.79 \\
 & EBK                  & 0.1526          & 44.5049         & 6.5581          & 3.4639          & 52.34 \\
 & M5P                  & -0.0204         & 51.1987         & 7.0752          & 3.5909          & 48.18 \\
\cmidrule(lr){2-7}
\multirow{4}{*}{Min GWL} 
 & Groundhog (Proposed) & \textbf{0.9624} & \textbf{1.3711} & \textbf{1.1710} & \textbf{0.7269} & \textbf{23.11} \\
 & IDW                  & 0.0225          & 42.6947         & 6.4075          & 2.9299          & 68.58 \\
 & EBK                  & 0.0318          & 42.0942         & 6.3692          & 3.2848          & 76.72 \\
 & M5P                  & -0.1924         & 48.5138         & 6.8785          & 3.3879          & 94.85 \\
\bottomrule
\end{tabular}
\end{table}

\begin{figure}[htbp]
    \centering
    \includegraphics[width=\linewidth]{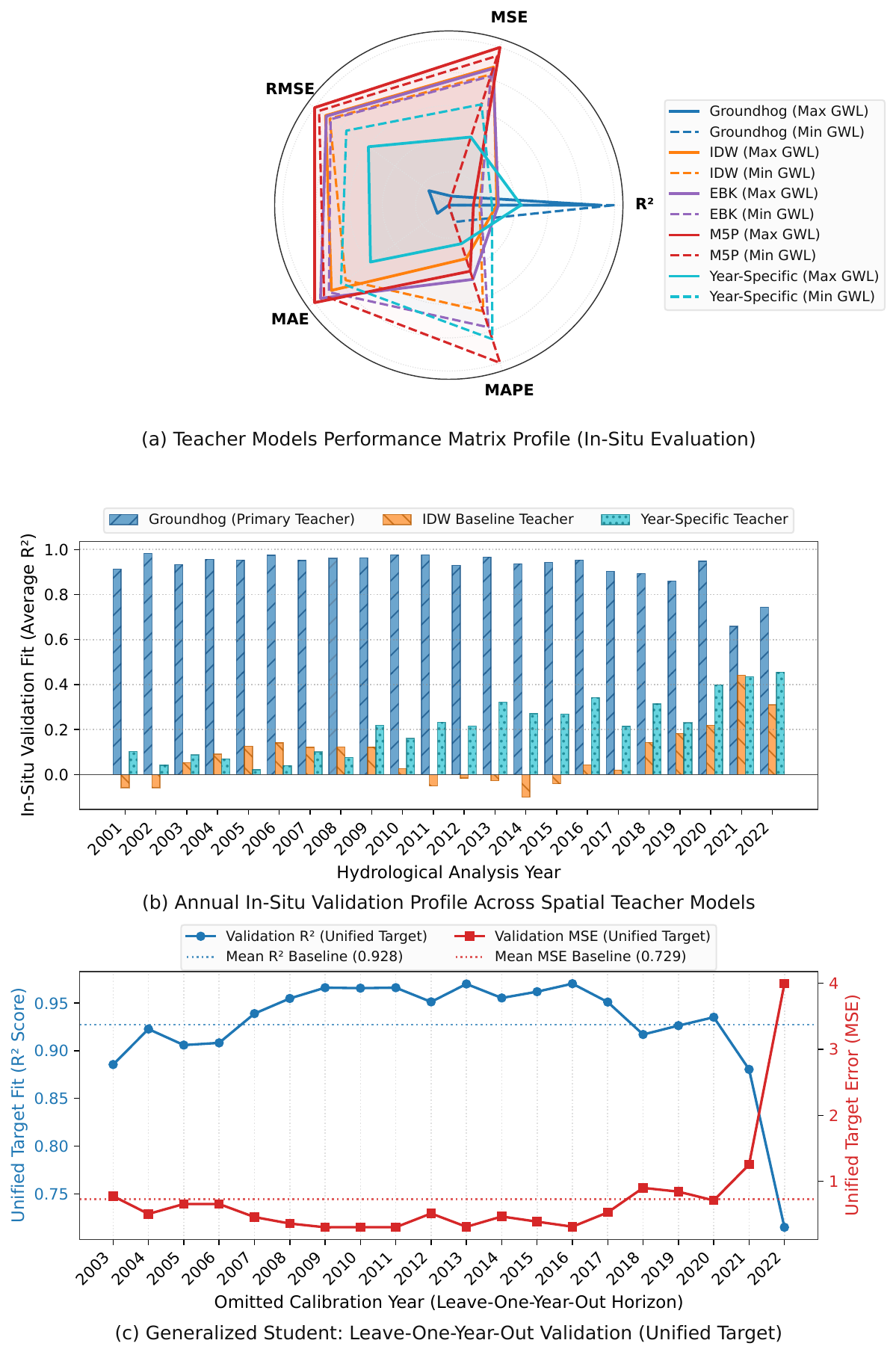}

    % \vspace{0.3cm} % Space between rows

    \caption{\textbf{Performance Metrics:}  
    \textbf{a.} Radar plot for different metrics of IDW, GroundHog and Year-Specific Random Forest.  \textbf{b.} $R^2$ scores of IDW, GroundHog and Year-Specific Random Forest.  
\textbf{c.} Leave-One-Year-Out (LOYO) validation for the Upsampling Model.   
}
    \label{fig:metrics}
\end{figure}
\subsection{Results from Task 1: Teacher model}

The key finding from Task 1 was that pooling observations across hydrological years substantially improved groundwater-level reconstruction (Table~\ref{tab:teacher_candidates_task1}). As shown in Panel A, year-specific training produced poor Random Forest performance for maximum GWL ($R^2=0.31$) and minimum GWL ($R^2=0.11$). In contrast, pooling observations across years in Panel B markedly improved performance for maximum GWL ($R^2=0.88$, MAE = 0.88~m) and minimum GWL ($R^2=0.97$, MAE = 0.70~m). This improvement was consistent across models, with Random Forest providing the strongest overall machine-learning performance and therefore serving as the backbone of GroundHog Teacher model.

The chosen Groundhog Teacher model also substantially outperformed conventional spatial interpolation when evaluated against held-out \textit{in-situ} observations, as shown in Panel C of Table~\ref{tab:teacher_candidates_task1}. For maximum GWL, Groundhog achieved $R^2=0.85$ (MSE = 4.19~m$^2$, MAE = 0.97~m), compared with IDW ($R^2=0.14$) and EBK ($R^2=0.15$). For minimum GWL, Groundhog achieved $R^2=0.96$ (MSE = 1.37~m$^2$, MAE = 0.73~m), whereas both IDW and EBK had $R^2<0.04$. M5P performed worse than the interpolation baselines for both seasonal extremes, with negative $R^2$ values. The radar plot further highlights Groundhog's consistent advantage across evaluation metrics (Fig.~\ref{fig:metrics}a).

This advantage was also evident across individual hydrological years (Fig.~\ref{fig:metrics}b). Groundhog achieved strong performance in most years, with $R^2>0.9$ in 18 of 22 years and $R^2>0.95$ in 13 years, while retaining useful performance even in 2021 ($R^2=0.66$) and 2022 ($R^2=0.74$). In contrast, IDW remained below $R^2=0.25$ through 2020 and reached only 0.44 and 0.31 in 2021 and 2022, respectively, while year-specific Random Forest performed substantially worse ($R^2=0.02$–$0.46$). Overall, pooling information across years produced more accurate and stable groundwater reconstructions than either year-specific modelling or conventional spatial interpolation.

\begin{table}[htbp]
\centering
\small
\caption{Performance Comparison of Task 2 Student Model Candidates Across Multi-Scale Upsampling Targets. \textbf{Bold} values indicate top-performing models, while \underline{underlined} values denote second-best performers.}
\label{tab:task2_student_models}
\begin{tabular}{llcccc}
\toprule
\textbf{Target Group} & \textbf{Metric} & \textbf{Random Forest} & \textbf{Neural Network} & \textbf{XGBoost} & \textbf{LightGBM} \\
\midrule
\multirow{4}{*}{Overall} 
 & $R^2$   & \textbf{0.9344} & 0.7125 & \underline{0.8355} & 0.7487 \\
 & MSE     & \textbf{0.6466} & 2.9228 & \underline{1.6891} & 2.6023 \\
 & RMSE    & \textbf{0.8041} & 1.7096 & \underline{1.2997} & 1.6132 \\
 & MAE     & \textbf{0.3769} & 1.0707 & \underline{0.7714} & 0.9432 \\
\cmidrule(lr){1-6}
\multirow{4}{*}{Min GWL} 
 & $R^2$   & \textbf{0.9191} & 0.6803 & \underline{0.8233} & 0.7383 \\
 & MSE     & \textbf{0.6322} & 2.4991 & \underline{1.3812} & 2.0452 \\
 & RMSE    & \textbf{0.7951} & 1.5808 & \underline{1.1752} & 1.4301 \\
 & MAE     & \textbf{0.3998} & 0.9557 & \underline{0.6803} & 0.8079 \\
\cmidrule(lr){1-6}
\multirow{4}{*}{Max GWL} 
 & $R^2$   & \textbf{0.9496} & 0.7448 & \underline{0.8477} & 0.7591 \\
 & MSE     & \textbf{0.6610} & 3.3468 & \underline{1.9972} & 3.1593 \\
 & RMSE    & \textbf{0.8130} & 1.8294 & \underline{1.4132} & 1.7775 \\
 & MAE     & \textbf{0.3541} & 1.1858 & \underline{0.8624} & 1.0786 \\
\bottomrule
\end{tabular}
\end{table}

\subsection{Results from Task 2: Student Model}
\label{subsec:result_task_2}

In Task 2 as well, Random Forest consistently provided the strongest Student model across all groundwater-level targets, while leave-one-year-out validation demonstrated robust performance across hydrological years. For the overall target ($R^2=0.93$, MAE = 0.38~m), minimum GWL ($R^2=0.92$, MAE = 0.40~m), and maximum GWL ($R^2=0.95$, MAE = 0.35~m), Random Forest substantially outperformed XGBoost, Neural Network, and LightGBM (Table~\ref{tab:task2_student_models}). These results motivated its selection as the Student model. The selected Random Forest maintained $R^2>0.9$ in 17 of 20 held-out years under leave-one-year-out (LOYO) validation, with a mean $R^2=0.93$ (Fig.~\ref{fig:metrics}c). Performance remained strong in 2021 ($R^2=0.88$, MSE = 1.25~m$^2$) and 2022 ($R^2=0.71$, MSE = 3.99~m$^2$), demonstrating strong temporal generalization overall.

\begin{figure}[!htbp]
    \centering
   
    \includegraphics[width=\textwidth]{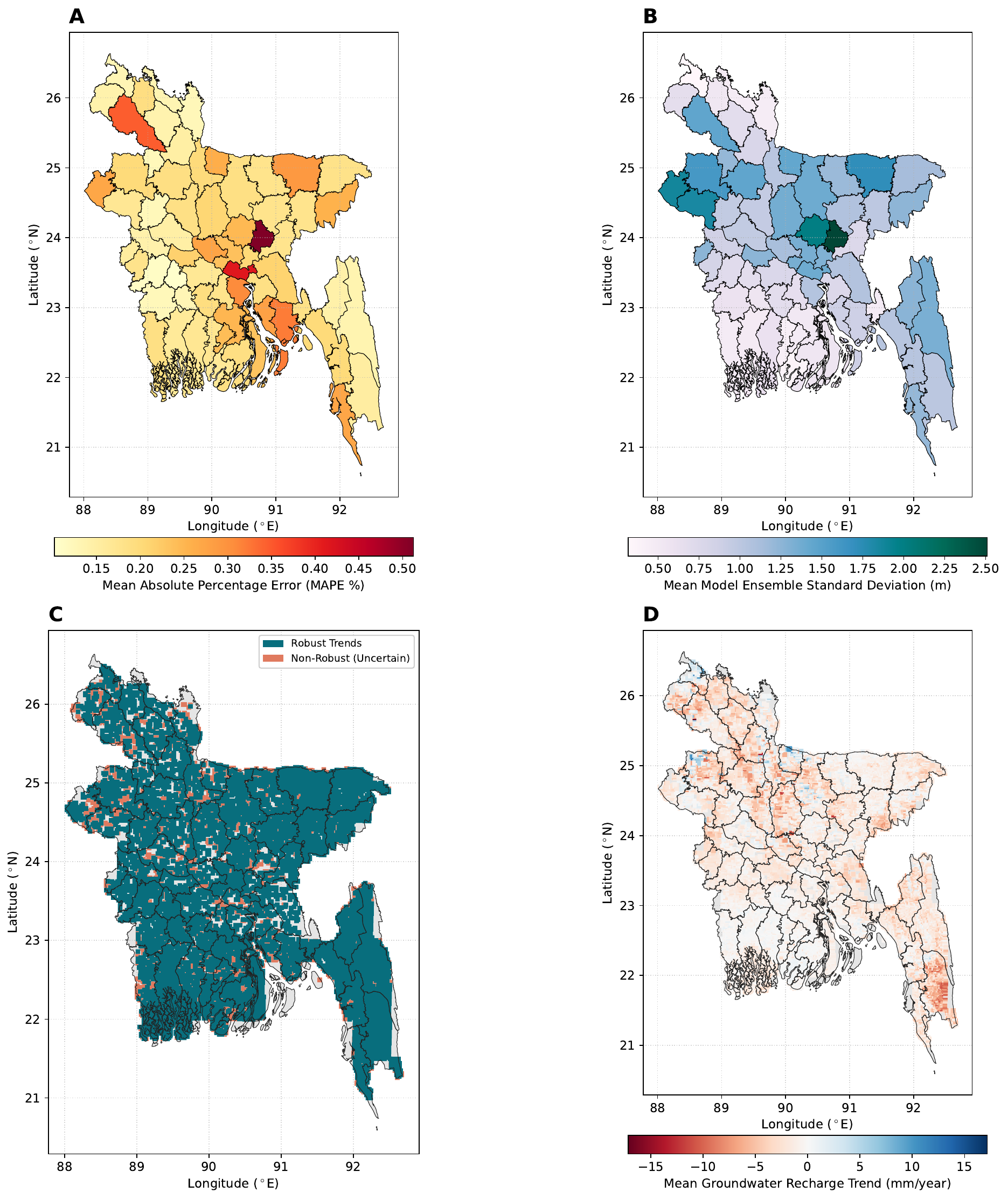}
    \caption{Geospatial model validation, epistemic uncertainty, and Monte Carlo trend robustness suite for groundwater recharge across Bangladesh. \textbf{(A)} District-level Leave-One-District-Out (LODO) cross-validation error, quantified via Mean Absolute Percentage Error (\text{MAPE} \%), highlighting regions where localized anthropogenic extraction or unique physical boundary conditions limit out-of-fold generalization. \textbf{(B)} Random Forest ensemble model uncertainty, measured as the mean standard deviation across individual decision trees ($m$). \textbf{(C)} Binary classification of point-wise trend robustness derived from $1,000$ Monte Carlo simulation iterations, isolating statistically robust trends (deep teal; empirical $95\%$ confidence interval strictly excluding zero) from non-robust signals (soft coral; error envelope spanning zero). \textbf{(D)} Point-wise mean annual groundwater recharge trend slopes (\text{mm/year}), using a symmetric diverging colormap to differentiate localized recharge accretion (blue) from temporal depletion (red).}
    \label{fig:lodo_uncertainty_robustness}

    \label{fig:lodo}
\end{figure}

Geographic generalization was assessed using Leave-One-District-Out (LODO) cross-validation. In each of 64 iterations, the model was trained on observations from 63 districts and tested on the remaining district, using observations from all available years to evaluate its ability to predict groundwater depth in an unseen region. Across all held-out districts, the Student model achieved an overall $R^2 = 0.70$ ($\text{RMSE} = 1.57~\text{m}$; $\text{MAPE} = 19.21\%$), demonstrating slightly better performance for maximum compared to minimum groundwater depth ($R^2 = 0.73$ versus $0.67$). Prediction errors varied spatially (Fig.~\ref{fig:lodo_uncertainty_robustness}A), with the highest MAPE concentrated in Narsingdi ($51.23\%$), followed by Munshiganj ($41.53\%$) and Dinajpur ($34.67\%$). In contrast, errors were minimized across much of western and southwestern Bangladesh, including Magura ($10.18\%$) and Jhenaidah ($11.12\%$) (Fig.~\ref{fig:lodo_reference_map}). To explicitly quantify spatial predictive uncertainty, we mapped the variance of predictions across all individual decision trees within the Random Forest ensemble of our Student model (Fig.~\ref{fig:lodo_uncertainty_robustness}B). This ensemble variability largely mirrored the LODO error distribution: it revealed higher epistemic uncertainty in complex regions like Narsingdi, while demonstrating high model confidence and stability across the southwestern coastal districts.

Furthermore, because our subsequent analysis examines recharge trends, we assessed whether these trends remain reliable when uncertainties in reconstructed groundwater depths and specific yield are taken into account. We conducted a 1,000-iteration Monte Carlo simulation (Detailed in Supplementary Section~\ref{sec:supp_monte_carlo}), perturbing predicted maximum and minimum groundwater depths using prediction-uncertainty estimates from LODO validation and varying specific yield in each iteration. We then recalculated annual recharge and its temporal trend at each location. A trend was considered robust when the lower and upper bounds of its 95\% simulation interval had the same sign (both positive or both negative), indicating that the simulated trends consistently supported the same direction. When the bounds had opposite signs, the trend direction was considered uncertain. By this criterion, 71.20\% of locations exhibited robust trends (Fig.~\ref{fig:lodo_uncertainty_robustness}C), including 63.92\% with robust negative trends and 7.25\% with robust positive trends (Fig.~\ref{fig:lodo_uncertainty_robustness}D). These results indicate that the reconstructed recharge-trend directions are robust to the uncertainties represented in the simulation across much of Bangladesh.

\begin{figure}[!htbp]
\centering
\includegraphics[width=\linewidth]{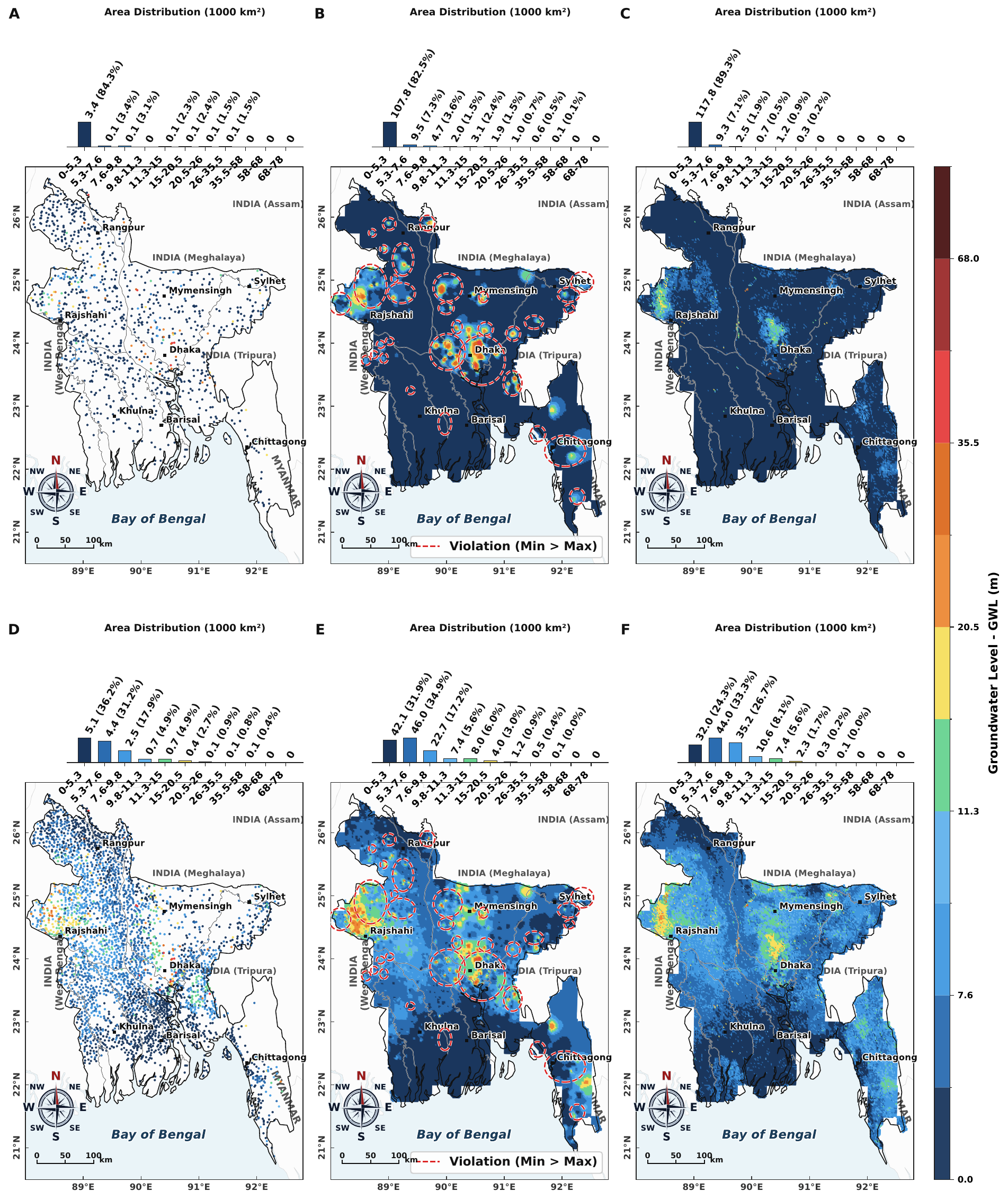}
\caption{\textbf{Comparison of sparse observations, IDW interpolation and pseudo-ground-truth groundwater-level maps for 2008.} Maps show depth below ground level (m), with maximum and minimum groundwater levels shown in the bottom and top rows, respectively. \textbf{(a,d)} Sparse \textit{in-situ} observations; \textbf{(b,e)} IDW interpolation, which produces physically inconsistent regions where minimum depth exceeds maximum depth (circled); and \textbf{(c,f)} GroundHog Teacher predictions, which provide continuous spatial fields while preserving the physical ordering of groundwater-level extremes.}
\label{fig:original_idw_pseudo}
\end{figure}

\begin{figure}[!htbp]
\centering
\includegraphics[width=0.8\linewidth]{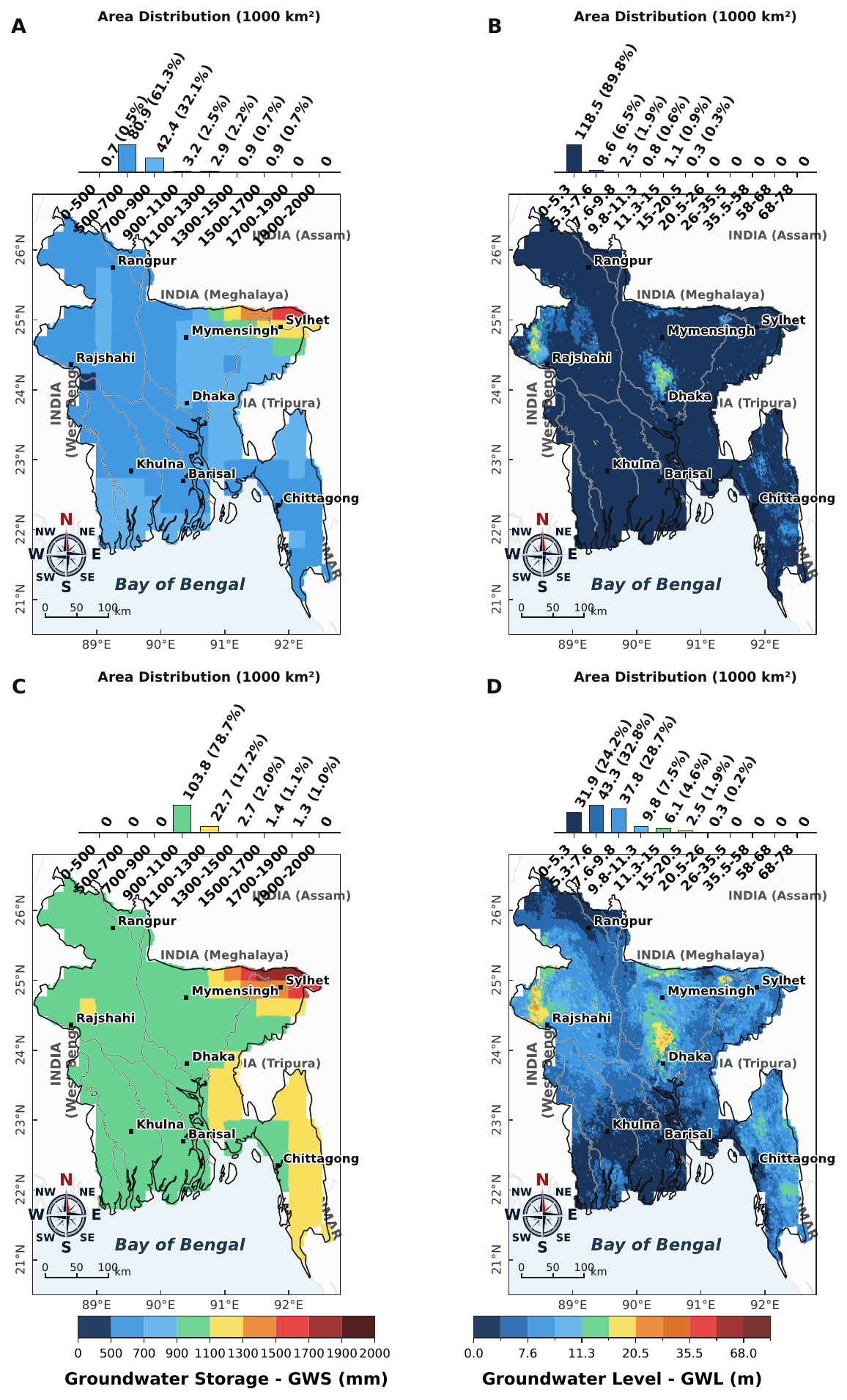}
\caption{\textbf{Task 2: GLDAS groundwater storage and upsampled groundwater levels for 2008.} \textbf{(a,c)} GLDAS minimum and maximum groundwater storage at 25~km resolution (mm); \textbf{(b,d)} corresponding upsampled annual minimum and maximum groundwater levels (m below ground level).}
\label{fig:gldas_upsampled}
\end{figure}

\subsection{Qualitative Analysis and Model Interpretation}

The spatial reconstructions show a clear contrast between conventional interpolation and the GroundHog Teacher model (Fig.~\ref{fig:original_idw_pseudo}). IDW produces overly smooth groundwater-depth patterns and misses important spatial details across years. It also produces areas where minimum groundwater depth exceeds maximum depth, which is physically impossible; the 2008 reconstruction provides a clear example (Fig.~\ref{fig:original_idw_pseudo}.b,e, circled). In contrast, the GroundHog Teacher recovers richer spatial patterns that agree with the sparse \textit{in-situ} observations while preserving the correct ordering of annual groundwater-depth extremes by conditioning minimum groundwater depth on the estimated maximum depth (Fig.~\ref{fig:original_idw_pseudo}.c,f). The Student model then uses these Teacher-derived fields to translate coarse ($25~\text{km}$) groundwater-storage estimates into fine-scale ($2~\text{km}$) groundwater-depth fields for annual minimum and maximum conditions (Fig.~\ref{fig:gldas_upsampled}). The resulting maps retain the spatial patterns learned by the Teacher across years~\cite{google_drive_folder}. The framework was also applied to the unseen year 2024; results are presented in Supplementary Section~\ref{sec:future_pred}.

To understand what drives the predictions, we examined feature importance, partial dependence plots (PDPs), and SHAP values (Supplementary Figs.~\ref{fig:pdp_and_corr}--~\ref{fig:upsample_min_shap}). The models relied primarily on hydrogeological characteristics, with elevation, clay thickness, lithology, and specific yield among the most influential predictors (Supplementary Figs.~\ref{fig:feature_and_shap}, ~\ref{fig:upsample_feat_imp}). The Student model also captured the expected physical relationship between groundwater storage and depth, with higher GLDAS groundwater storage associated with shallower groundwater and lower storage with deeper groundwater (Supplementary Figs.~\ref{fig:upsample_max_shap},~\ref{fig:upsample_min_shap}). Detailed feature-importance, PDP, correlation, and SHAP results are provided in the Supplementary Information.

\begin{figure}[!htbp]
    \centering
   \includegraphics[width=\linewidth]{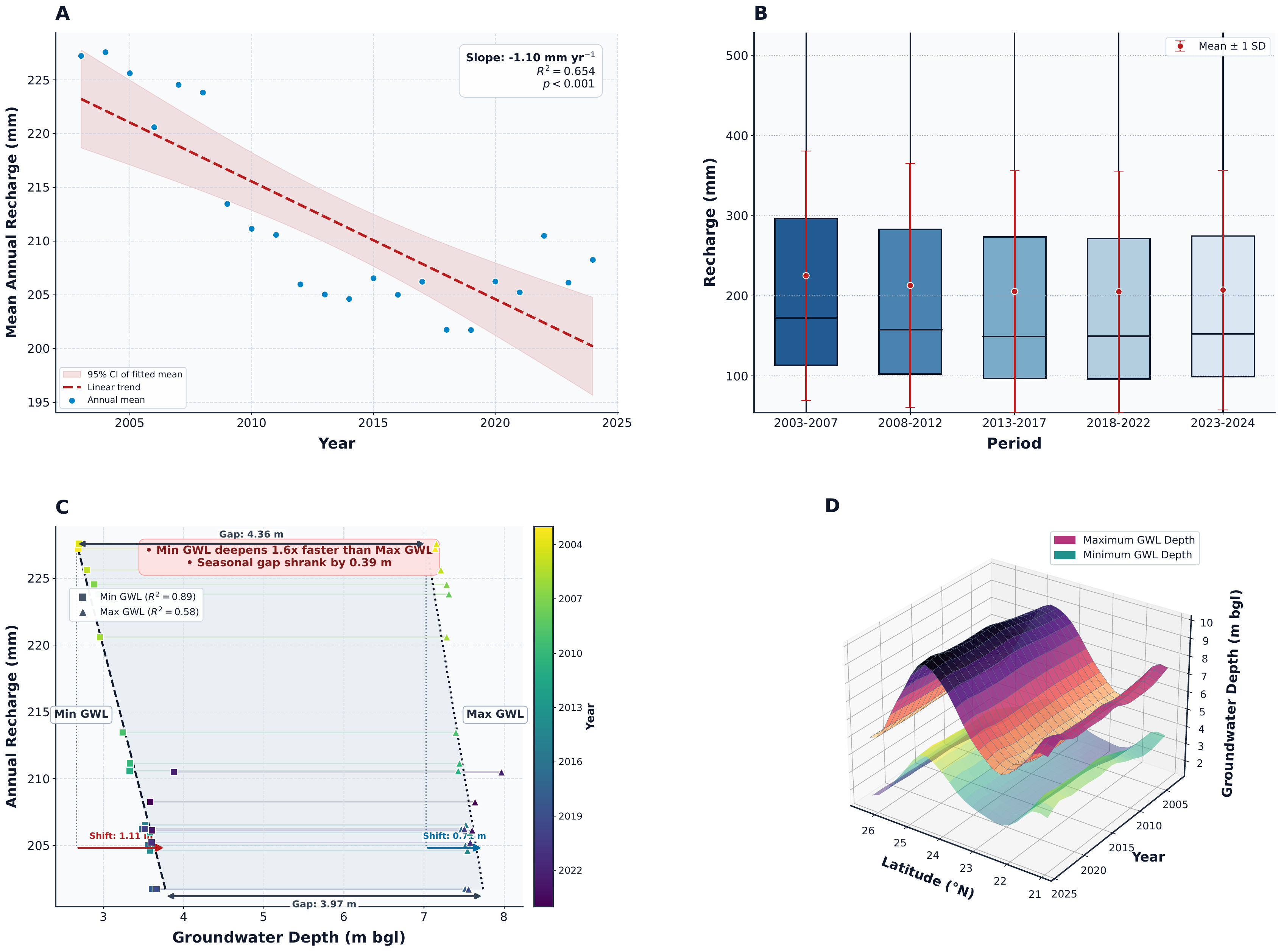}
\caption{Spatiotemporal changes in national-average groundwater recharge and water-table depth, 2003--2024. \textbf{(A)} Linear trend in national-average annual recharge, showing a statistically significant decline of 0.110~cm~yr$^{-1}$ ($R^2=0.654$, $p<0.001$); the 95\% confidence interval is shaded. \textbf{(B)} Distribution of annual recharge across 5-year epochs; boxes show interquartile ranges and red points show means $\pm$ 1 standard deviation. \textbf{(C)} Trajectory of national-average annual recharge (mm) against seasonal groundwater-level extremes (m bgl), with points colored by year. Regression bounds and dimension arrows illustrate a differential deepening effect: the annual minimum (shallower) groundwater level deepens at a faster rate than the maximum, leading to a progressively shrinking seasonal fluctuation gap. \textbf{(D)} Latitudinal variation in groundwater depth through time.}
\label{fig:gwl_recharge_dynamics}
\end{figure}

\subsection{National-scale decline in groundwater recharge and water-table levels}

The record reconstructed from our estimates reveals a sustained decline in groundwater replenishment across Bangladesh between 2003 and 2024. National mean annual recharge declined from 227.2~mm in 2003 to 208.3~mm in 2024, a reduction of 19.0~mm (8.4\%). This decline was statistically significant ($-1.10$~mm~yr$^{-1}$, $p < 0.001$; Fig.~\ref{fig:gwl_recharge_dynamics}A), with the overall distribution of annual recharge shifting progressively downward (Fig.~\ref{fig:gwl_recharge_dynamics}B). These national-scale results are complemented by a Monte Carlo robustness analysis of local recharge trends (Sec.~\ref{subsec:result_task_2}), which propagated LODO-derived groundwater-depth prediction uncertainties and specific-yield variability. Together, these analyses provide complementary perspectives on recharge change: the national trend captures changes in annual mean recharge across Bangladesh, while the Monte Carlo analysis assesses the direction and robustness of trends at individual grid locations. The mean simulated trend across locations was $-1.4963 \pm 1.9983$~mm~yr$^{-1}$, with 63.92\% of locations exhibiting robust negative trends and 7.25\% exhibiting robust positive trends. Locations with robust negative trends had a mean trend of $-1.9617$~mm~yr$^{-1}$ (Fig.~\ref{fig:lodo_uncertainty_robustness}C,D).

The decline in recharge coincided with progressive water-table deepening, fundamentally altering the system's seasonal dynamics (Fig.~\ref{fig:gwl_recharge_dynamics}C). As annual recharge dropped, the entire system trajectory shifted deeper into the aquifer. Notably, this deepening was highly asymmetric: mean minimum (shallower/wet-season) groundwater depth increased by 90~cm (from 2.68~m to 3.58~m), while mean maximum (deeper/dry-season) depth increased by 50~cm (from 7.14~m to 7.64~m). Both trends were statistically significant ($p < 0.001$). Because the minimum depth deepened nearly twice as fast as the maximum depth ($4.84$ versus $2.44$~cm~yr$^{-1}$), the seasonal fluctuation gap progressively compressed over the 21-year period (Fig.~\ref{fig:gwl_recharge_dynamics}C).

Beneath these national averages lies substantial spatial variation, with latitudinal profiles revealing the most severe water-table deepening in central to north-central Bangladesh, where minimum groundwater depth increased by 168~cm and maximum depth by 118~cm (Fig.~\ref{fig:gwl_recharge_dynamics}D). While the mean difference of 3.82~m between minimum (wet-season) and maximum (dry-season) water levels highlights the system's massive seasonal fluctuation, the pronounced long-term deepening indicates that monsoonal recovery is increasingly failing to offset dry-season drawdowns in these highly impacted regions.

\begin{figure}[!htbp]
    \centering
    \includegraphics[width=\linewidth]{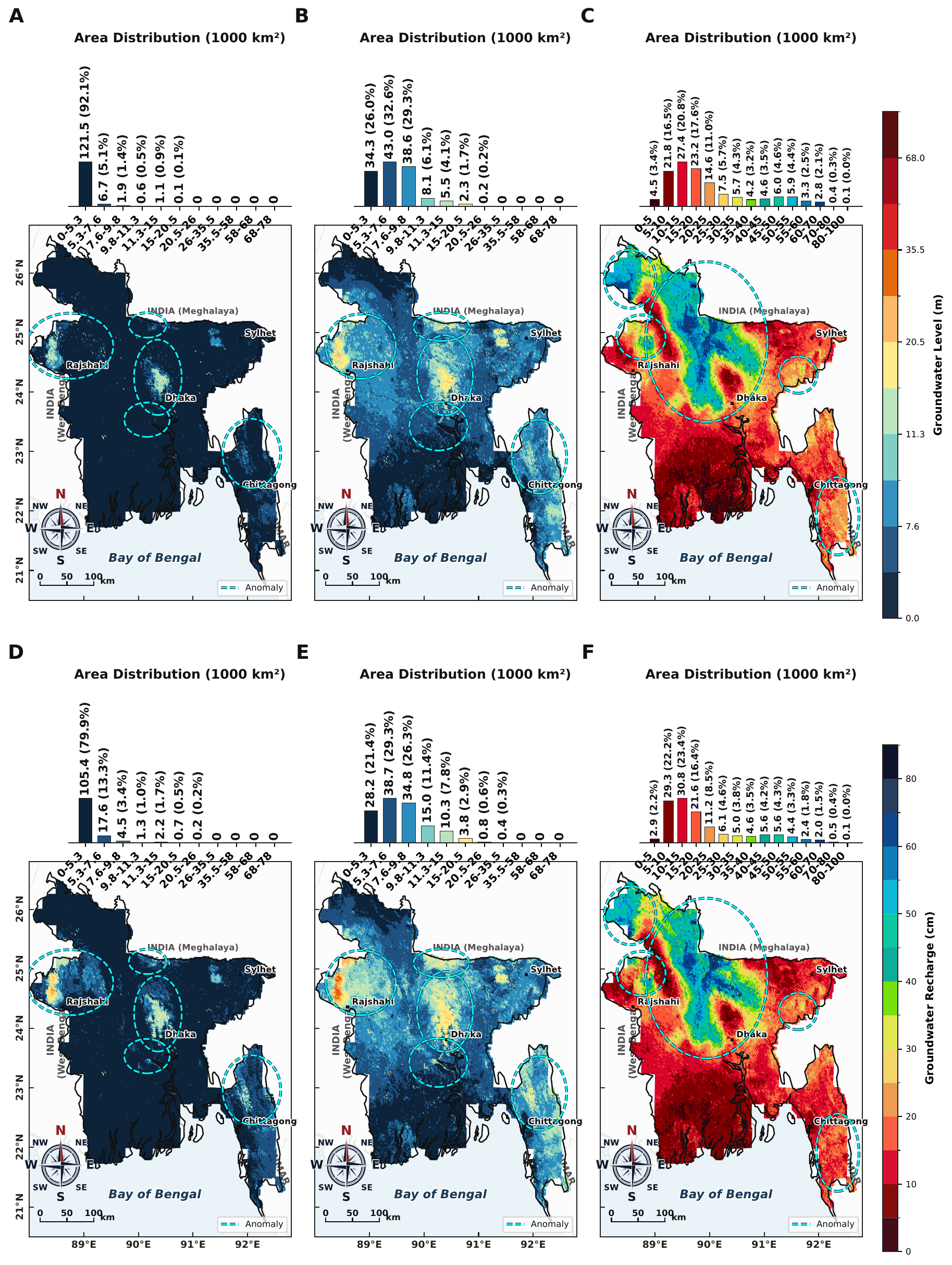}
    \caption{\textbf{Temporal comparison of upsampled results:} Spatial distribution of upsampled values in 2003 (top row) and 2022 (bottom row) for \textbf{a,b.} annual minimum groundwater depth (m bgl), \textbf{c,d.} annual maximum groundwater depth (m bgl), and \textbf{e,f.} annual recharge (cm).}
    \label{fig:temporal}

\end{figure}

\subsection{Spatial patterns of groundwater decline and recharge change}
Spatially, groundwater decline between 2003 and 2022 was most pronounced around Dhaka, the northwestern Barind Tract and Tista Floodplain, and the southeastern Hill Tracts (Fig.~\ref{fig:temporal}). In 2003, 90.8\% of Bangladesh maintained shallow annual minimum groundwater depths (0--5.3~m below ground level). By 2022, the water table in the northwest and Dhaka deepened significantly, reaching up to 43~m at 2-km grid centers and 73~m at individual \textit{in-situ} monitoring locations. Annual maximum groundwater depth also deepened over this period, though by a smaller average of 0.44~m. These spatial trajectories align with documented depletion driven by intensive irrigation, rapid urbanization, and heavy groundwater abstraction \cite{moshfika2022assessing,nowreen2017groundwater,hoque2007declining,roy2021assessment,hossain2019mar,shahrearassessment}.

Recharge declined across all analyzed physiographic regions overall, with the largest average reductions occurring in the Tista Floodplain (mean change of -36.3~mm, with localized point reductions up to 440.7~mm) and the Barind Tract (mean change of -23.6~mm, with point reductions up to 176.5~mm). Substantial localized declines were also recorded in the Brahmaputra Floodplain (up to 308.7~mm), the Chittagong \& Eastern Hills (up to 161.7~mm), and the Madhupur Tract (up to 141.4~mm). Positive regional anomalies were rare; most notably, Sherpur district within the Brahmaputra Floodplain deviated from the broader regional pattern by registering a net positive mean recharge shift (+21.3~mm) (see Fig.~\ref{fig:ae_reference_map} for mentioned region locations). Areas of declining recharge coincided with groundwater deepening, particularly across the northwest and densely developed zones, indicating persistent regional depletion consistent with previous assessments \cite{parvin2019unsustainable,zzaman2021application,nowreen2017groundwater,moshfika2022assessing,tabassum2026Investigating}.

\begin{figure}[!htbp]
    \centering
    \includegraphics[width=\linewidth]{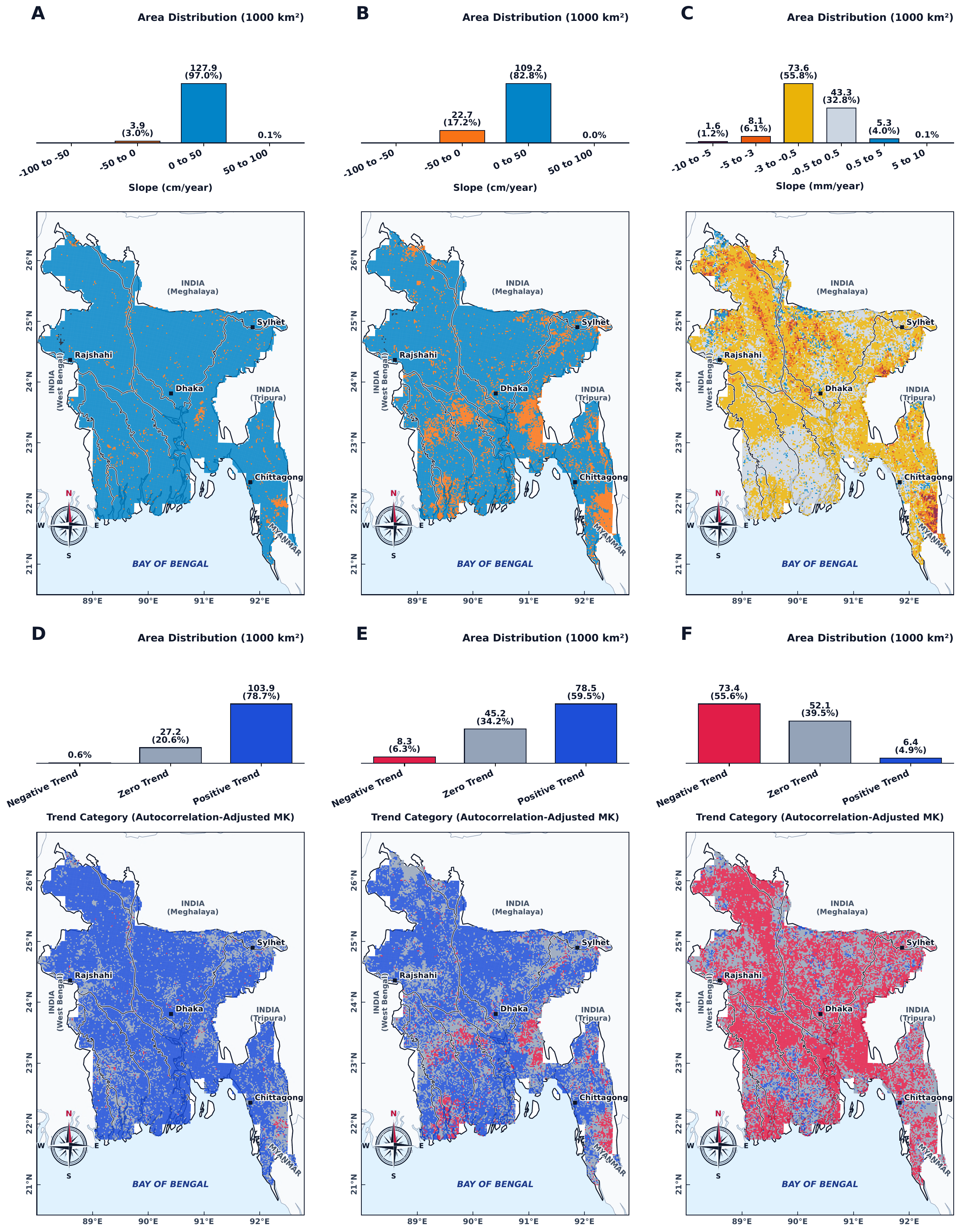}
    \caption{\textbf{Spatial trends in groundwater levels and recharge:} Sen's slope (top row; mm~yr$^{-1}$ for recharge, cm~yr$^{-1}$ for groundwater depth) and autocorrelation-adjusted Modified Mann--Kendall significance ($p^* < 0.05$; bottom row) for \textbf{a,d.} minimum groundwater depth, \textbf{b,e.} maximum groundwater depth, and \textbf{c,f.} recharge. For recharge, Sen's slopes are grouped into six categories: negligible trends ($-0.5$ to $0.5$~mm~yr$^{-1}$; light grey), moderate decline ($-3$ to $-0.5$~mm~yr$^{-1}$; yellow), strong decline ($-5$ to $-3$~mm~yr$^{-1}$; orange), severe decline ($-10$ to $-5$~mm~yr$^{-1}$; dark red), and positive accretion ($0.5$ to $10$~mm~yr$^{-1}$; blue shades).}
    \label{fig:trends}
\end{figure}

\subsection{Spatial trends in groundwater levels and recharge}
Trend analysis shows that groundwater decline and recharge loss were widespread across Bangladesh rather than confined to differences between selected years (Fig.~\ref{fig:trends}). Minimum groundwater depth showed a positive Sen's slope across most of the country (98.2\%), with standard Mann--Kendall (MK) tests indicating positive trends across 87.1\% of the study area (Supplementary Fig.~\ref{fig:supp_standard_mk}A). After accounting for interannual temporal autocorrelation via the Modified Mann--Kendall (MMK) test (Detailed in Sec.~\ref{sec:recharge_trend}), 78.7\% retained statistical significance (Fig.~\ref{fig:trends}D). Maximum groundwater depth similarly showed a positive Sen's slope across 88.6\% of the nation, with positive MK trends across 63.6\% (Supplementary Fig.~\ref{fig:supp_standard_mk}B) adjusting to 59.5\% under the MMK framework (Fig.~\ref{fig:trends}E). Because increasing groundwater depth below ground level (bgl) indicates a deeper water table, these persistent positive trends point to widespread national groundwater depletion.

Recharge showed a similarly widespread decline across Bangladesh. Standard MK analysis initially identified negative recharge trends across 65.7\% of the country and positive trends across 5.3\% (Supplementary Fig.~\ref{fig:supp_standard_mk}C). When controlling for lag-$k$ serial correlation using the MMK test, a dominant majority of 55.6\% retained statistically significant recharge decline, whereas significant positive accretion occurred across only 4.9\% (Fig.~\ref{fig:trends}F). Based on Sen's slope, 54.0\% of the country experienced \textit{declining recharge} ($-3$ to $-0.5~\text{mm~yr}^{-1}$), while a further 6.9\% showed \textit{strong declines} exceeding $-3~\text{mm~yr}^{-1}$ (Fig.~\ref{fig:trends}C). Another 34.6\% showed \textit{little or no change} ($-0.5$ to $0.5~\text{mm~yr}^{-1}$), whereas only 4.4\% showed \textit{increasing recharge} exceeding $0.5~\text{mm~yr}^{-1}$. The strongest recharge declines occurred around Dhaka, in the northwestern Tista floodplains along with the Barind region, and across parts of the southeastern Hill Tracts, whereas positive trends were largely confined to localized corridors along the Brahmaputra floodplain (see Fig.~\ref{fig:ae_reference_map} for the locations of these regions).

\section{Discussion}

Across Bangladesh, groundwater levels are deepening in tandem with declining recharge, signaling a widespread weakening of aquifer replenishment. While water-table declines do not automatically equate to unsustainable depletion, their persistent co-occurrence with predominantly negative recharge trends points to diminishing groundwater availability. This pattern is particularly pronounced around Dhaka, the Tista floodplain, the Barind Tract, and parts of the southeastern Hill Tracts, where escalating water demand coincides with hydrogeological conditions that limit natural recharge. Our generalizable framework makes these patterns observable across space and time by translating coarse, time-varying groundwater storage into \(2~\text{km}\) groundwater-depth maps, enabling high-resolution recharge reconstruction over more than two decades.

The spatial pattern also provides a more nuanced perspective on the Bengal Water Machine (BWM) hypothesis. Previous work proposed that groundwater pumping can induce substantial additional recharge across parts of Bangladesh \cite{shamsudduha2022bengal}. In contrast, our reconstruction demonstrates that statistically significant negative recharge trends dominate 55.6\% of the country after controlling for temporal autocorrelation (compared with 65.7\% under standard Mann--Kendall testing), while statistically significant positive trends occur across only 4.9\% (5.3\% under standard testing). This disparity deepens when considering the magnitude of change: only 4.4\% of the country exhibits increasing recharge exceeding \(0.5~\text{mm yr}^{-1}\) based on Sen's slope. By either criterion, increasing recharge is concentrated mainly along localized corridors of the Brahmaputra floodplain. This does not negate the BWM mechanism; rather, it indicates that its expression is spatially constrained to favorable hydrogeological settings. Pumping-induced recharge may therefore offer local resilience where surface-water connectivity permits, but its limited spatial footprint dictates that it cannot serve as a broad national safeguard against groundwater depletion.

These findings carry direct implications for groundwater management. Regions experiencing persistent water-table deepening alongside declining recharge represent emerging groundwater-stress hotspots, demanding proactive efforts to curb demand and evaluate recharge enhancement strategies. Where suitable surface-water sources and hydrogeological conditions align, Managed Aquifer Recharge (MAR) offers a viable strategy for augmenting natural replenishment \cite{hossain2019mar,usgs1999groundwater}. Conversely, areas exhibiting resilient or increasing recharge—such as the Brahmaputra floodplain—rely on critical surface-water and hydrogeological connectivity that must be protected from environmental degradation. Rather than dictating a uniform national response, the spatial heterogeneity revealed by our reconstruction strongly supports geographically differentiated assessments of groundwater stress and mitigation potential.

Although our framework significantly improves the spatiotemporal coverage of groundwater estimates, several limitations remain. The Student model inherently carries residual uncertainty from the Teacher-generated pseudo-ground-truth fields; these fields are modeled estimates and should not be interpreted as direct observations. Additionally, the framework does not explicitly parameterize spatial changes in groundwater extraction associated with population growth, industrial expansion, or land-use shifts. Incorporating dynamic abstraction data could further improve the attribution of observed groundwater changes to specific anthropogenic drivers. Finally, applying this framework beyond Bangladesh requires calibration and validation against observations representative of the target hydrogeological setting. While these limitations must be considered when interpreting highly localized changes in light of our uncertainty analysis, they do not compromise the broad, national-scale trends identified here.

Ultimately, our reconstruction reveals that Bangladesh is experiencing widespread groundwater deepening compounded by declining recharge, with enhanced replenishment restricted to localized corridors. While the Bengal Water Machine provides vital regional buffering, its limited footprint underscores that it is not a panacea for national groundwater security. By seamlessly downscaling coarse satellite data into fine-scale annual depth maps that generalize across unseen years and districts, our Teacher--Student framework resolves these complex dynamics. Because it generates high-resolution estimates from accessible environmental data without requiring continuous new field measurements, this approach establishes a highly scalable foundation for identifying emerging water stress and guiding sustainable management in data-sparse aquifers worldwide.

\section{Materials and Methods}

\subsection{Data preprocessing}

Groundwater-level observations were compiled from the Bangladesh Water Development Board (BWDB), Bangladesh Agricultural Development Corporation (BADC), Department of Public Health Engineering (DPHE), and Barind Multipurpose Development Authority (BMDA). The dataset contains annual maximum and minimum groundwater levels (m bgl) from 17,821 sites during 2001--2022. The observations are spatially and temporally sparse, with approximately 1,000 sites reporting measurements in a typical year. Across the study period, 59,443 maximum and 26,813 minimum groundwater-level observations were available, including 26,670 site-year pairs with both measurements (Fig.~\ref{fig:distribution}, Supplementary Information). We also compiled GLDAS groundwater storage and environmental variables at approximately 25~km resolution for 200 grids during 2003--2024 \cite{grace_gldas}. All variables were aligned by year and spatial location before model training. For conventional model evaluation, 20\% of observations from each year were held out for testing, with the remaining 80\% used for training. Generalization across time and space was evaluated separately using Leave-One-Year-Out (LOYO) and Leave-One-District-Out (LODO) cross-validation. Detailed spatial preprocessing and construction of the model-ready datasets are provided in Supplementary Methods, Section~\ref{sec:geospatial_preprocessing}.

\subsection{Features Used}
\label{sec:features_main}

The framework uses 17 hydrogeological features (HGFs), which are grouped according to how quickly they vary over time. Fifteen features are treated as static and describe the underlying physical setting of each location, including topography, drainage, soil and subsurface properties. These include elevation, slope, aspect, curvature, profile curvature, plan curvature, drainage density, distance from stream, topographic wetness index (TWI), terrain ruggedness index (TRI), stream topographic index (STI), stream power index (SPI), specific yield ($S_y$), lithology, and lithological clay thickness. Two features, annual mean NDVI and NDWI, are treated as slowly varying and capture changes in surface vegetation and water conditions over time.

The framework also uses highly dynamic groundwater-storage (GWS) estimates from GLDAS. These annual maximum and minimum GWS estimates are available at $25~\text{km}$ resolution and provide the time-varying groundwater-storage signal used by the Student model. During downscaling, the 17 HGFs are represented at both the regional $25~\text{km}$ GLDAS-grid scale and the local $2~\text{km}$ target scale. Thus, the model combines relatively stable hydrogeological characteristics with slowly changing surface conditions and time-varying groundwater storage to reconstruct groundwater depth at $2~\text{km}$ resolution. Detailed feature definitions and source datasets are provided in Table~\ref{table:features} of the Supplementary Materials, while spatial aggregation and categorical-feature preprocessing are described in Supplementary Methods, Section~\ref{sec:geospatial_preprocessing}.

\subsection{Task 1: Teacher Model and Pseudo-Ground Truth}
\label{sec:task1_teacher}

The Teacher generates high-resolution annual groundwater-depth supervision from sparse \textit{in-situ} observations. All available observations from 2001--2022 are pooled into a single multi-year dataset, with year included as a numerical predictor alongside the 17 HGFs to capture interannual variability. Two separate Random Forest models are used to reconstruct annual maximum and minimum groundwater depths. The minimum-depth model uses the same inputs as the maximum-depth model, with the predicted maximum depth included as an additional conditioning variable to prevent physically inconsistent estimates in which the minimum exceeds the maximum. During training, the observed maximum depth was used; during inference, the maximum depth predicted by the Phase 1 model was used.

During inference, the trained models are applied to the HGFs and year of 36,096 points at $2~\text{km}$ resolution from the ArcGIS grid to generate dense annual maximum and minimum groundwater-depth maps. These predictions constitute the ``Pseudo-Ground Truth''. To avoid assigning pseudo-ground-truth values to locations that are already represented by \textit{in-situ} observations, predictions within 1.85~km of an \textit{in-situ} observation are excluded. This distance is slightly smaller than the nominal $2~\text{km}$ grid spacing, allowing some grid points to be retained between closely spaced observations while avoiding overlap with observed locations. The remaining predictions are combined with the original \textit{in-situ} observations to form the supervisory dataset for Task 2. Model implementation details are provided in Supplementary Methods, Section~\ref{sec:model_implementation}.

\subsection{Task 2: Student Model}
\label{sec:task2_student}

GroundHog, the Student model, learns to downscale annual $25~\text{km}$ GLDAS GWS into $2~\text{km}$ groundwater depth. Each target location is represented by three complementary inputs: regional GWS, regional HGFs describing the parent $25~\text{km}$ GLDAS grid, and local HGFs describing the target $2~\text{km}$ location. The regional features provide the broader hydrogeological context, whereas the local features capture fine-scale spatial variability. The Student is trained using the combined \textit{in-situ} and Teacher-generated supervisory dataset and predicts annual maximum and minimum groundwater depths directly at $2~\text{km}$ resolution. Geographic identifiers are excluded from training to promote spatial generalization. The construction of the Student input dataset is detailed in Supplementary Methods, Section~\ref{sec:geospatial_preprocessing}.

\subsection{Tasks 3--4: Recharge Estimation and Trend Analysis}
\label{sec:recharge_trend}

For Task 3, annual recharge for 2003--2024 was estimated from the reconstructed maximum and minimum groundwater depths using the Water Table Fluctuation method:

\begin{equation}
\text{Recharge} =
(\text{Maximum groundwater depth}
-\text{Minimum groundwater depth})
\times S_y \times 1000,
\label{eq:recharge}
\end{equation}

where $S_y$ is aquifer-specific specific yield. The resulting recharge estimates are expressed in millimetres. Subsequently, Task 4 assessed long-term trends in annual maximum and minimum groundwater depth and recharge over 2003--2024 using Sen's slope estimator \cite{sen1968estimates} and the conventional Mann--Kendall trend test \cite{MannKendallTest}. 

To account for temporal autocorrelation (serial dependence) in annual hydrogeological time series, which can exaggerate the statistical significance of trends, we applied the Modified Mann--Kendall (MMK) test \cite{hamed1998modified, yue2002modified} (Detailed in Supplementary Sec~\ref{sec:supp_mmk}). The test first removes the linear trend and then calculates lag-$k$ autocorrelation coefficients from the detrended residuals. When significant autocorrelation was detected ($p^* < 0.05$), a correction factor was applied to adjust the variance of the test statistic, $V(S)$. This correction accounts for the persistence of values between years and provides a more reliable assessment of long-term trends.

\subsection{Evaluation Metrics and Validation Strategy}
\label{sec:metrics_used}

Model performance was evaluated using the coefficient of determination ($R^2$), mean squared error (MSE), mean absolute error (MAE), root mean squared error (RMSE), and mean absolute percentage error (MAPE). We evaluated model generalization across both time and space using Leave-One-Year-Out (LOYO) and Leave-One-District-Out (LODO) cross-validation. In LOYO, one year is held out for testing and the model is trained on all other years; this is repeated for each year. In LODO, all observations from one district are held out for testing, while the model is trained on observations from all other districts. LOYO therefore tests whether the model can predict groundwater conditions in an unseen year, whereas LODO tests whether it can predict groundwater conditions in a district not used for training.

% \begin{figure}
%     \centering
%     \includegraphics[width=\linewidth]{NatureWater_Aesthetics_3x2_2003_vs_2022_CyanRings_HighContrast.pdf}
%     \caption{Caption}
%     \label{fig:placeholder}
% \end{figure}

%%%%%%%%%%%%%%%% SUPPLEMENTARY TEXT %%%%%%%%%%%%%%%

\FloatBarrier
\backmatter

\section*{Declarations}
\subsection*{Competing interests}
The authors declare that there are no competing interests.

\subsection*{Funding}
This research work was supported by a RISE Internal Research Grant with ID 2022-01-025. Nowreen and Rahman are partially supported by Basic Research Grants of BUET.
\section*{Code and Data Availability}

Groundwater-level observations were obtained from the Bangladesh Water Development Board (BWDB), Bangladesh Agricultural Development Corporation (BADC), Department of Public Health Engineering (DPHE), and Barind Multipurpose Development Authority (BMDA). GLDAS-derived groundwater-storage and environmental data were also used \cite{grace_gldas}. Details of the datasets, preprocessing, and model development are provided in the Methods and Supplementary Information.

All data, trained model weights, preprocessing objects, and code required to reproduce the analysis are publicly available:

\begin{itemize}
\item \textbf{In-situ observations and 2~km hydrogeological features:} \url{https://www.kaggle.com/datasets/sakibahmed91/input-of-groundhog-psuedo-generator}
\item \textbf{GLDAS data:} \url{https://www.kaggle.com/datasets/sakibahmed91/gldas-fixed-data}
\item \textbf{Model weights, imputer, and scalers:} \url{https://www.kaggle.com/datasets/sakibahmed91/test-wights-for-groundhog}
\item \textbf{Notebook for Pseudo-ground-truth generation:} \url{https://www.kaggle.com/code/sakibahmed91/groundhog-psuedo-ground-truth-generator-public/edit/run/229337311}
\item \textbf{Notebook for Upsampling model training:} \url{https://www.kaggle.com/code/sakibahmed91/groundhog-for-training-upsampling-model}
\item \textbf{Notebook for Downstream analysis:} \url{https://www.kaggle.com/code/sakibahmed91/groundhog-downstream-task}
\item \textbf{GitHub repository:} \url{https://github.com/bojack-horseman91/GroundHog}
\end{itemize}

All models were trained using Kaggle notebooks without GPU acceleration.

\bibliography{sn-bibliography}% common bib file

\newpage
\clearpage

% --- Supplementary Information Title ---
\begin{center}
    \huge \textbf{Supplementary Information}
\end{center}
\vspace{1em}

%%%%%%%%%%%%%%%% START OF SUPPLEMENT %%%%%%%%%%%%%%%

% Figures, tables, equations and pages in the supplement are numbered S1, S2 etc.
% Prefix all supplementary numbering with S
\renewcommand{\thesection}{S\arabic{section}}
\renewcommand{\thefigure}{S\arabic{figure}}
\renewcommand{\thetable}{S\arabic{table}}
\renewcommand{\theequation}{S\arabic{equation}}
\renewcommand{\thepage}{S\arabic{page}}

% Reset all counters for the supplementary section
\setcounter{section}{1}
\setcounter{figure}{1}
\setcounter{table}{1}
\setcounter{equation}{1}
\setcounter{page}{1} % not 0, as new page already started
% References continue the numbering from the main text.
\begin{figure}
    \centering
    \includegraphics[width=\linewidth]{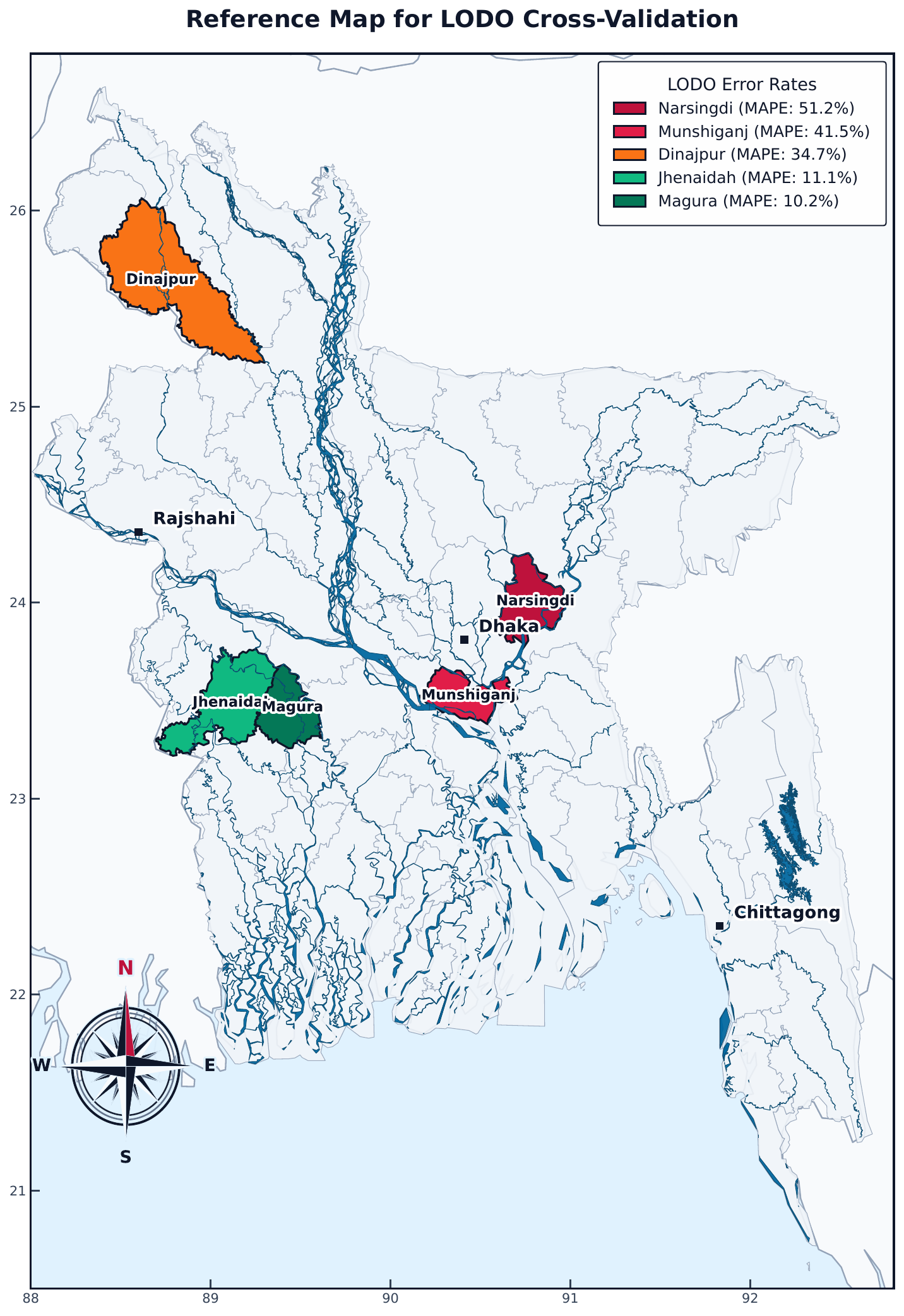}
    \caption{Geographic reference map of selected Bangladesh districts evaluated during Leave-One-District-Out (LODO) cross-validation.}
    \label{fig:lodo_reference_map}
\end{figure}

\begin{figure}
    \centering
    \includegraphics[width=\linewidth]{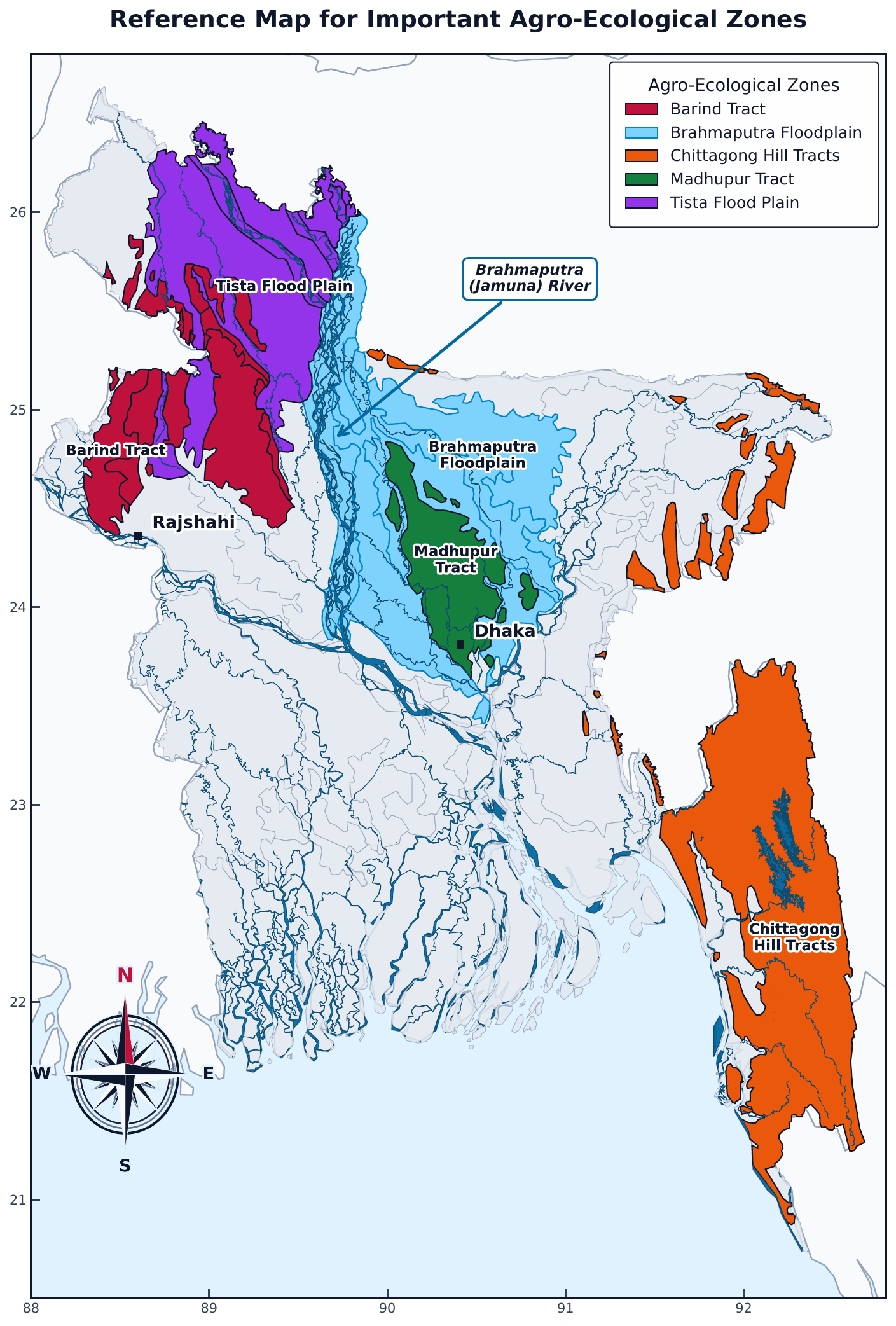}
    \caption{Distribution of key Agro-Ecological Zones (AEZs) and major river networks across Bangladesh.}
    \label{fig:ae_reference_map}
\end{figure}

\section{Detailed Model Interpretation}
\label{sec:interpretation}

\begin{figure}[!htbp]
    % \begin{subfigure}{\textwidth}
    %     \centering
    %     \includegraphics[width=0.7\linewidth]{Selected_max_gwl_feature_pdp.pdf}
    %     \caption{}
    %     \label{fig:PDP_maxGWL}
    % \end{subfigure}
    % \hfill
    % \begin{subfigure}{\textwidth}
    %     \centering
    %     \includegraphics[width=0.7\linewidth]{correlation_heatmap.png}
    %     \caption{}
    %     \label{fig:heatmap}
    % \end{subfigure}
    % \hfill
    % \begin{subfigure}{\textwidth}
    %     \centering
    %     \includegraphics[width=0.7\linewidth]{3D_SPI_vs_TWI.jpg}
    %     \caption{}
    %     \label{fig:PDP_spi_vs_twi}
    % \end{subfigure}
    \includegraphics[width=\linewidth]{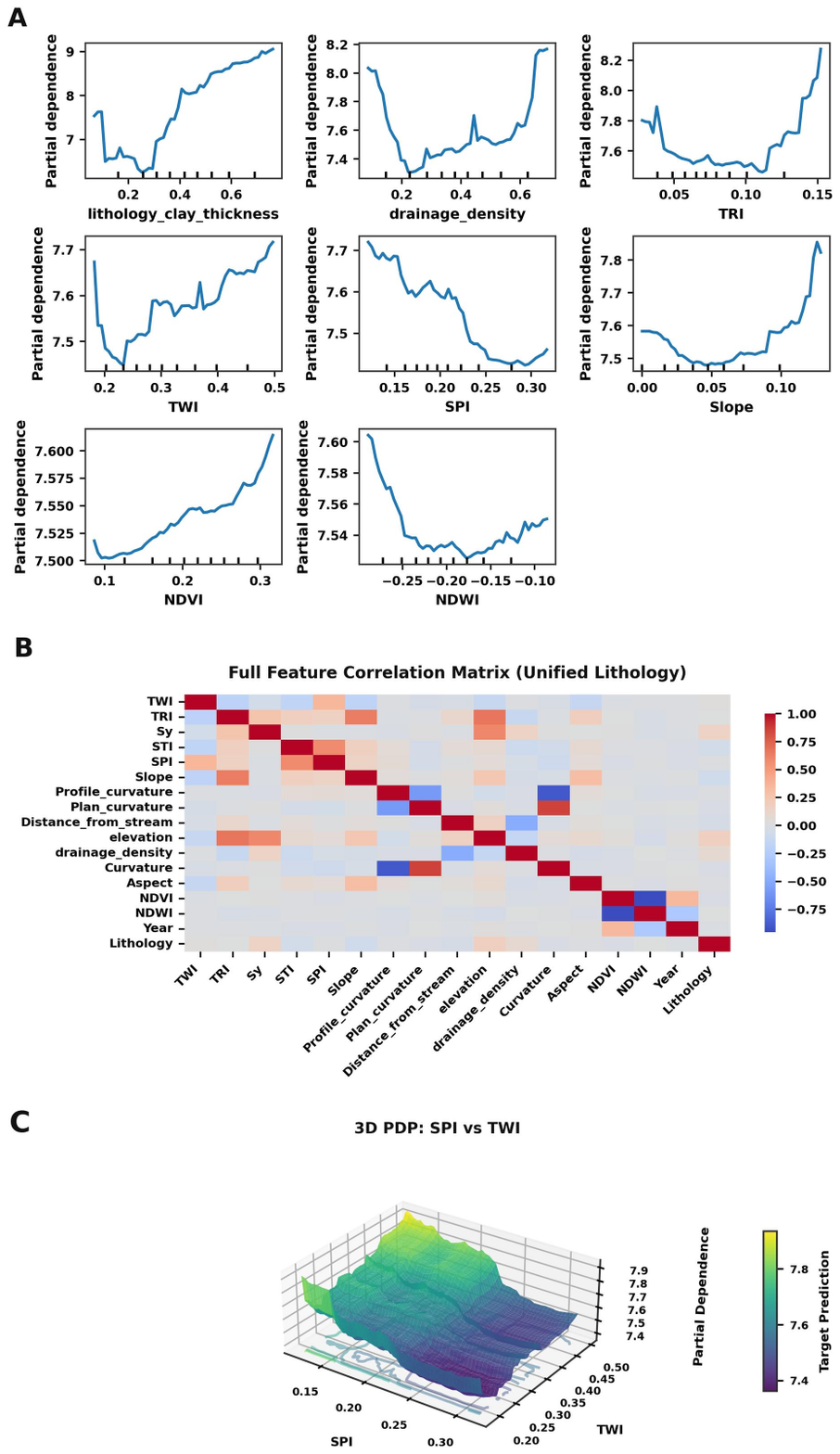}
 \caption{\textbf{Effects of Variable Changes on Groundwater Level:} \textbf{a.} Partial Dependence Plot (PDP) for the maximum groundwater level regressor illustrates feature impacts while holding others constant.  
\textbf{b.} Feature correlation heatmap highlights key relationships. 
\textbf{c.} 3D Partial Dependence Plot of Stream Power Index (SPI) and Topographic Wetness Index (TWI).}
    \label{fig:pdp_and_corr}
\end{figure}
\begin{figure}[!htbp]
    \centering
    \includegraphics[width=\linewidth]{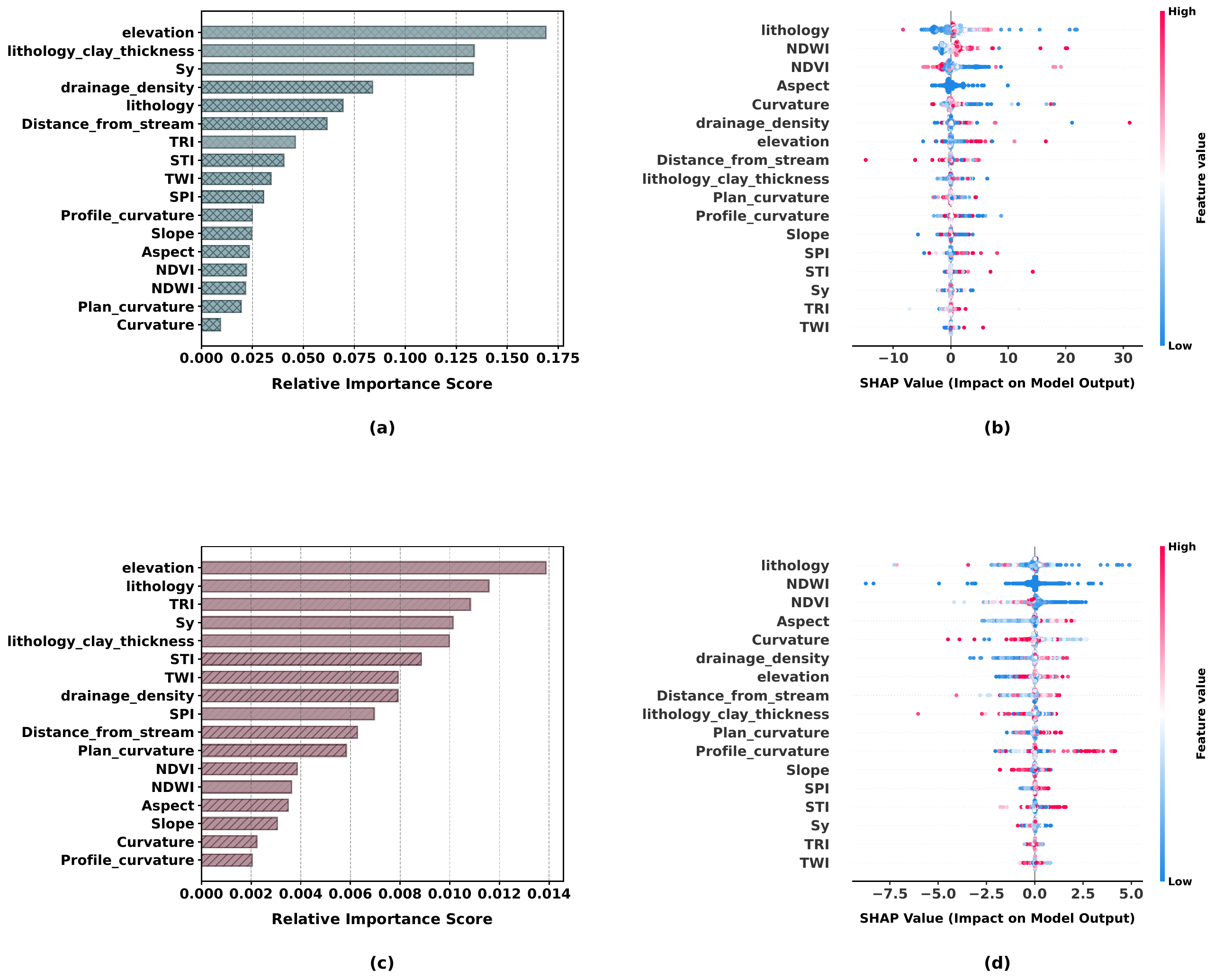}
   \caption{\textbf{Feature importance and SHAP values:} \small In the SHAP plot, red values indicate higher feature values, while blue values indicate lower feature values. The X-axis represents SHAP values, showing positive (high) or negative (low) influence on the output. \textbf{a, c} show feature importance for the Max and Min GWL Models, respectively. \textbf{b, d} present SHAP values for the Max and Min GWL Models, respectively.}
    \label{fig:feature_and_shap}
\end{figure}
\subsection{Partial Dependence Analysis}

Partial dependence plots (PDPs) were used to examine how individual predictors influence groundwater-depth predictions while holding other predictors constant. Because groundwater depth is measured below ground level (GWL bgl), higher predicted values indicate a deeper water table. The PDPs for the maximum GWL model (Fig.~\ref{fig:pdp_and_corr}A) reveal several relationships consistent with known controls on groundwater recharge and storage.

Increasing lithological clay thickness is associated with a deeper predicted water table. Thick clay layers can restrict vertical diffuse recharge and hydraulic exchange, providing a plausible explanation for this relationship \cite{Nowreen2020}. Drainage density shows a nonlinear relationship: the predicted water table is relatively deep at very low drainage density, becomes substantially shallower around 0.2, and then deepens again with further increases. The initial shallowing may reflect increased hydraulic connectivity and seepage from surface-water channels, whereas higher drainage density may promote runoff and reduce infiltration \cite{arulbalaji2019gis,anderson2010geomorphology}.

Terrain and hydroclimatic variables show similarly interpretable relationships. Increasing Terrain Ruggedness Index (TRI) and slope are associated with deeper groundwater, consistent with greater runoff and reduced infiltration on rugged and steep terrain \cite{verma2024spatial,gao2024comprehensive,han2020effect}. Lower Standardized Precipitation Index (SPI), indicating drier conditions, is also associated with deeper groundwater, consistent with groundwater losses during prolonged dry periods \cite{leelaruban2017examining}. Conversely, higher Normalized Difference Water Index (NDWI) is associated with shallower groundwater, consistent with greater surface-water availability and recharge \cite{arif2024using,antonsson2023space}.

The TWI relationship illustrates the importance of considering interactions among correlated predictors. When TWI is varied independently while SPI and other predictors are held constant, increasing TWI is associated with a deeper predicted water table, apparently contrasting with the conventional interpretation of TWI as an indicator of wetter terrain and shallower groundwater \cite{sorensen2006calculation}. However, TWI and SPI are correlated and jointly influence groundwater conditions. When both increase together, the two-dimensional PDP (Fig.~\ref{fig:pdp_and_corr}C) shows substantially shallower groundwater under wetter conditions, consistent with enhanced recharge \cite{purnama2016water,maqsoom2022delineating}. Thus, the joint response provides a more physically meaningful interpretation than the isolated TWI effect.

The feature-correlation heatmap (Fig.~\ref{fig:pdp_and_corr}B) further shows that NDWI and NDVI are strongly negatively correlated, while SPI and TWI are positively correlated. These relationships highlight the coupling among environmental predictors and reinforce the need to interpret individual PDPs in the context of correlated variables.

\subsection{Feature Importance and SHAP Analysis}

\begin{figure}
   \centering
        \includegraphics[width=\textwidth]{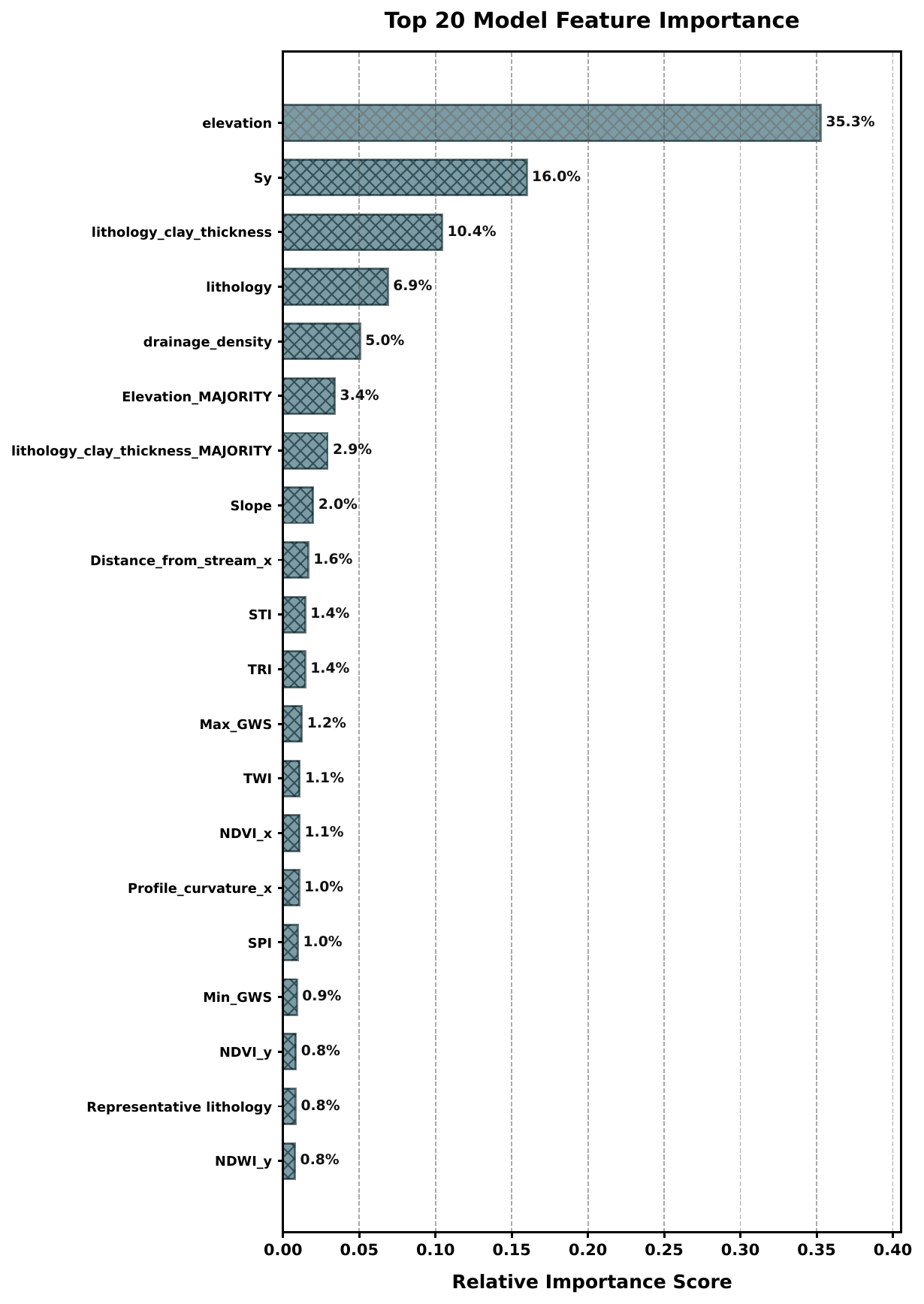}
        \caption{Feature Importance of Student Upsampling model}
        \label{fig:upsample_feat_imp}
\end{figure}

\begin{figure}
        \centering
        \includegraphics[width=\textwidth]{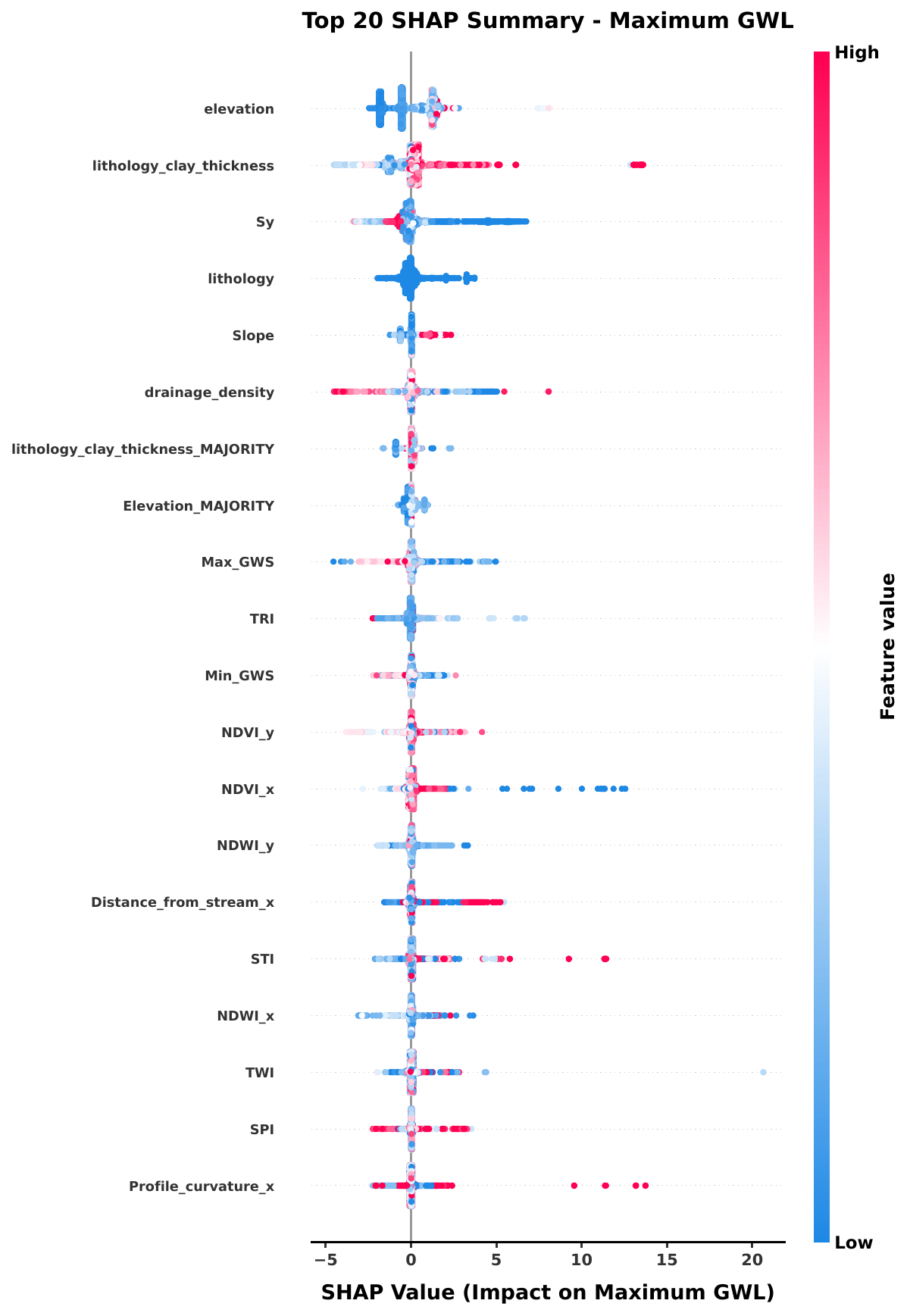}
        \caption{SHAP score of the Max GWL output of Student Upsampling model}
        \label{fig:upsample_max_shap}
    \end{figure}

\begin{figure}
        \centering
        \includegraphics[width=\textwidth]{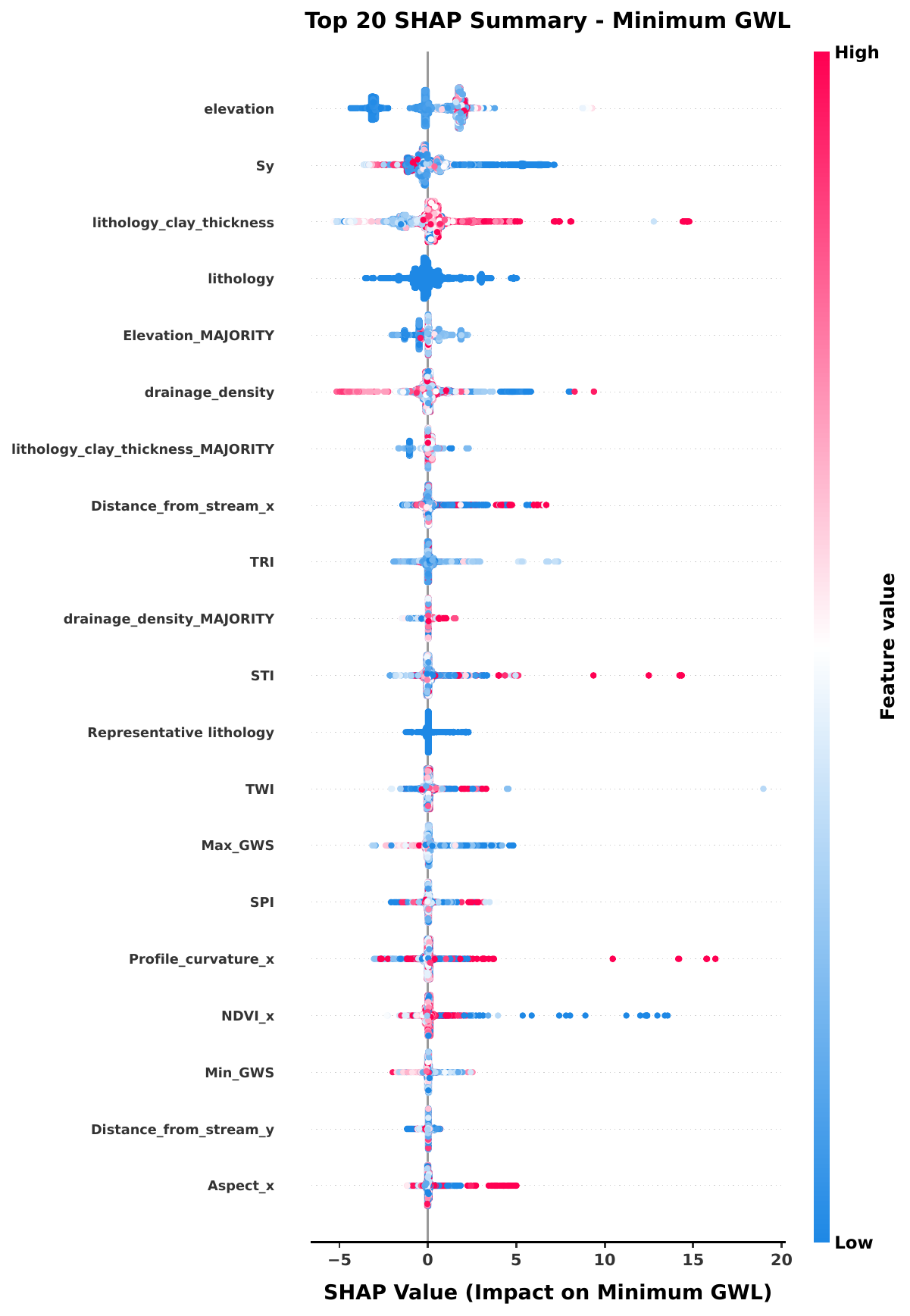}
        \caption{SHAP score of the Min GWL output of Student Upsampling model}
        \label{fig:upsample_min_shap}
    \end{figure}

Feature importance was calculated from the Random Forest models, while SHAP values were obtained using TreeExplainer to quantify the direction and magnitude of individual feature contributions \cite{van2022tractability,shap_treeexplainer}. Positive SHAP values indicate that a feature shifts the prediction toward a deeper groundwater table, whereas negative values shift it toward a shallower table. Feature-importance and SHAP results for the Teacher models are shown in Fig.~\ref{fig:feature_and_shap}; corresponding results for the Upsampling Student are provided in Figs.~\ref{fig:upsample_feat_imp}, \ref{fig:upsample_max_shap}, and \ref{fig:upsample_min_shap}.

Elevation was the most influential feature in both Teacher models and remained highly influential in the Upsampling Student, consistent with the role of topography in controlling groundwater gradients, recharge, and hydraulic connectivity \cite{zeng2022groundwater}. Lithological clay thickness, lithology, and specific yield were also among the dominant predictors, consistent with their roles in water retention and groundwater storage \cite{wang2018short,lin2023estimating}.

SHAP values further clarify these relationships. Higher clay thickness values generally produce positive SHAP values in the maximum GWL model, shifting predictions toward deeper groundwater, with similar patterns observed for the minimum GWL and Upsampling Student models (Fig.~\ref{fig:feature_and_shap}; Figs.~\ref{fig:upsample_max_shap}, \ref{fig:upsample_min_shap}).

The Upsampling Student also shows a clear relationship between GLDAS groundwater storage and predicted groundwater depth. Higher groundwater-storage values are generally associated with negative SHAP values and shallower predictions, whereas lower storage values produce positive SHAP values and deeper predictions. This provides a physically intuitive link between regional groundwater-storage variability and local water-table depth, supporting the Student model's ability to translate coarse, time-varying storage information into fine-scale groundwater-depth patterns.

Feature-importance and SHAP rankings are broadly consistent within the Teacher models. Differences for the Upsampling Student arise because feature importance is calculated for the combined model, whereas SHAP values are calculated separately for maximum and minimum groundwater-depth outputs. Together, the PDP, feature-importance, and SHAP analyses show that the models capture interpretable relationships among topography, hydroclimate, lithology, and groundwater storage.
 
\begin{figure}
    \centering
    \includegraphics[width=0.9\linewidth]{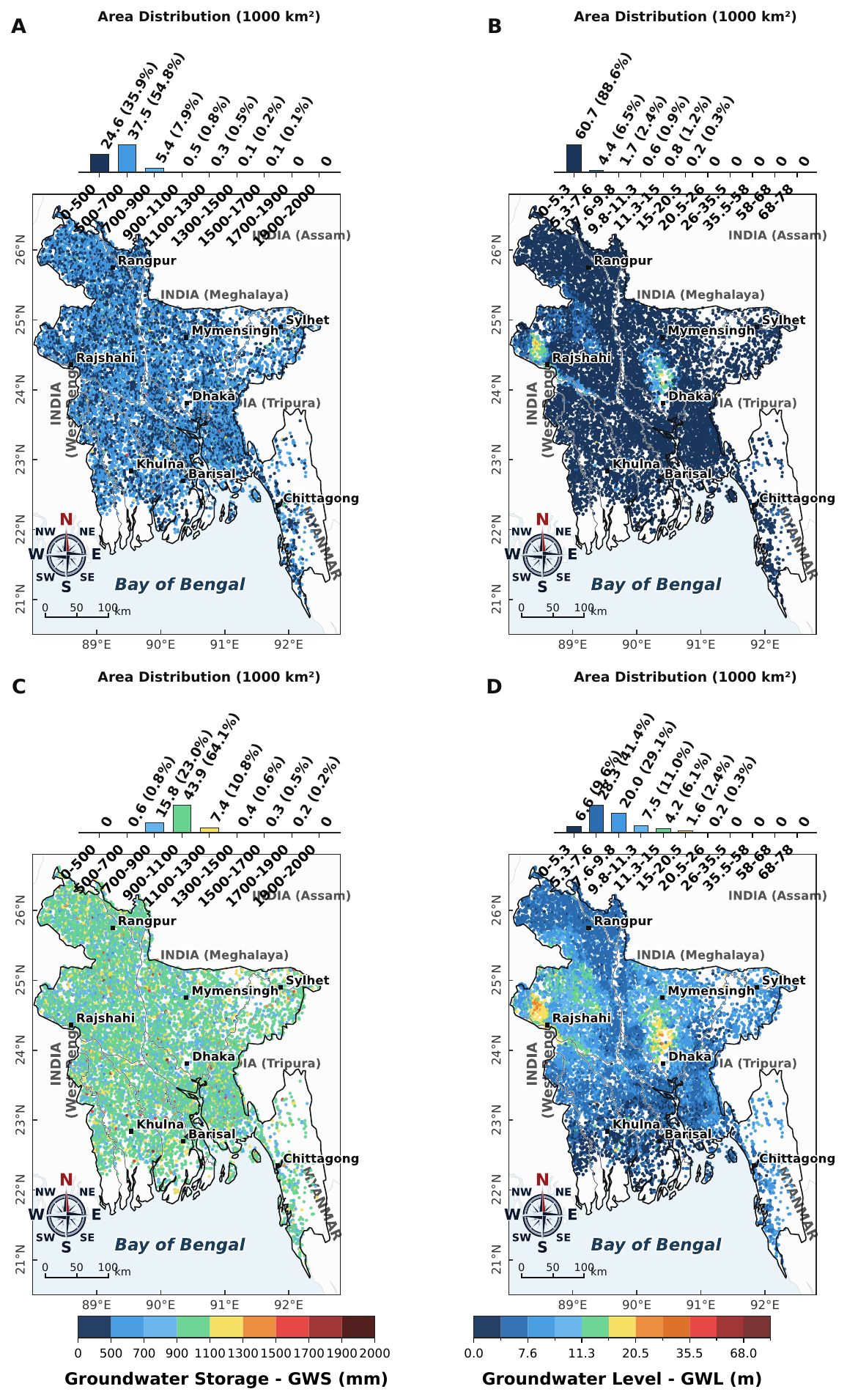}
    \caption{\textbf{Future Prediction for 2024:}  
    Leveraging our Upsampling Model, we have upsampled the 2024 GLDAS groundwater storage data to obtain high-resolution groundwater level estimates even before in-situ measurements become available. The upsampled data is shown for sites where various organizations in Bangladesh monitor groundwater levels.}
    \label{fig:future_pred}
\end{figure}

\section{Operational Forecasting: 2024 Groundwater Reconstruction}
\label{sec:future_pred}

A practical advantage of \textit{GroundHog} is that it can reconstruct groundwater conditions without waiting for nationwide \textit{in-situ} observations. Although automated loggers are increasingly used, much of Bangladesh's groundwater monitoring network still relies on manual data collection and entry, creating substantial delays before complete annual observations become available. This limits the timely assessment of changing groundwater conditions.

To demonstrate the operational capability of the framework, we used GLDAS data available through October 2024 to generate a continuous $2~\text{km}$ groundwater-depth map for 2024 (Fig.~\ref{fig:future_pred}). At the time of the analysis, nationwide official \textit{in-situ} observations for 2024 were not yet available. The reconstruction therefore demonstrates how \textit{GroundHog} can provide preliminary, spatially continuous groundwater information before conventional monitoring data are compiled, rather than representing a retrospectively validated 2024 estimate.

\begin{figure}
    \centering
    \includegraphics[width=\linewidth]{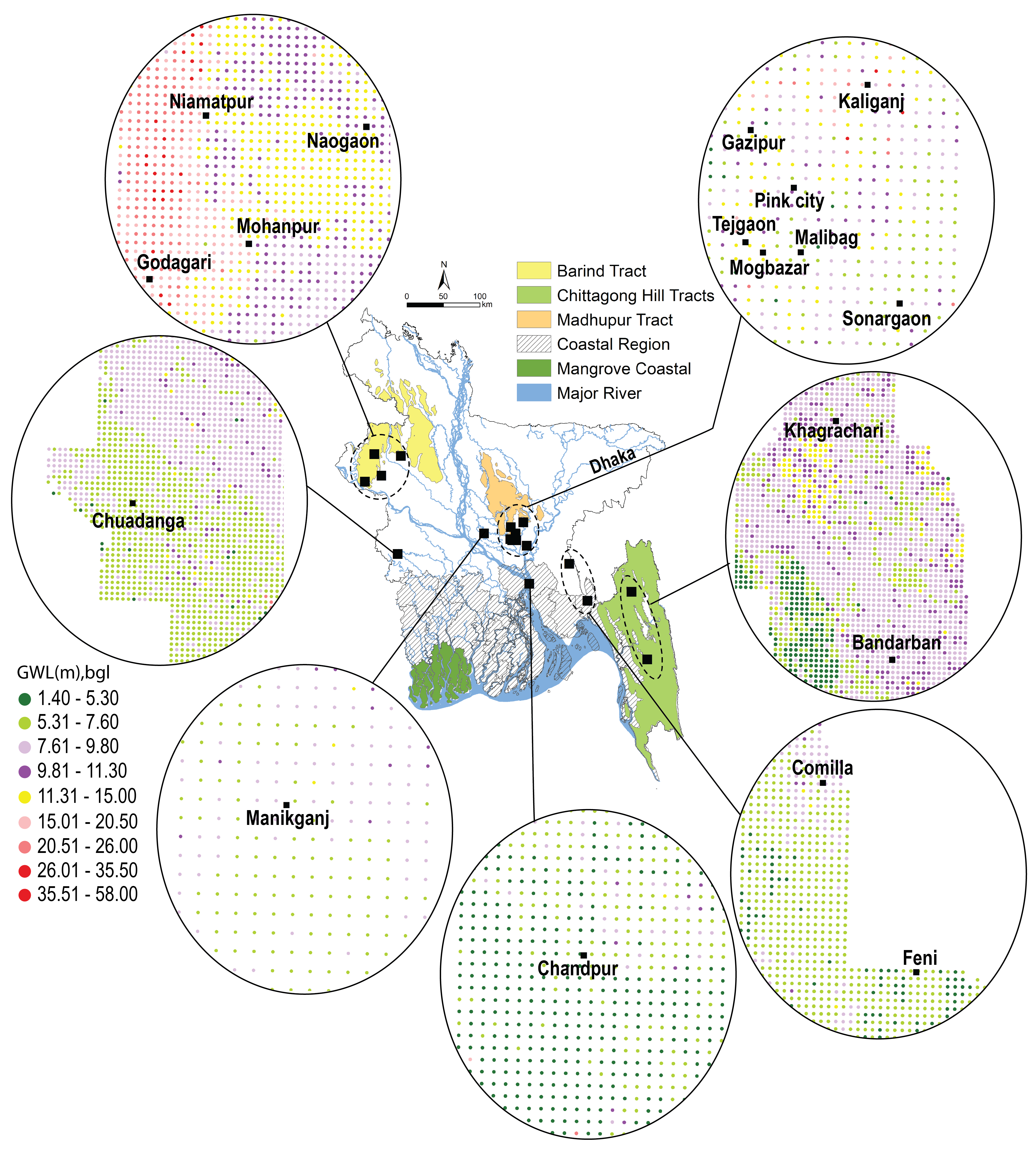}
    \caption{\textbf{Reference Map:} This map highlights various locations in Bangladesh that are mentioned in the manuscript.}
    \label{fig:reference_map}
\end{figure}

The 2024 reconstruction identifies pronounced groundwater depths in and around Dhaka, with values exceeding 55--60~m below ground level in highly urbanized areas such as Pink City, Malibagh, and Tejgaon, and reaching approximately 60~m or more in parts of Kaliganj (Gazipur) and Sonargaon (Narayanganj) (Fig.~\ref{fig:reference_map}). It also highlights rapid deepening around Dhaka. In Manikganj and its surrounding region, reconstructed groundwater depth increased from approximately 41~m in 2003 to 68~m in 2024, corresponding to a Sen's slope of $0.76~\text{m~yr}^{-1}$. These results illustrate how the framework can identify areas of pronounced groundwater stress for targeted monitoring and subsequent verification as new observations become available.

Because nationwide 2024 \textit{in-situ} observations are not currently available to us, these results should be interpreted as model-based reconstructions rather than independently validated measurements.

\begin{figure}[!htbp]
    \centering
    \includegraphics[width=\linewidth]{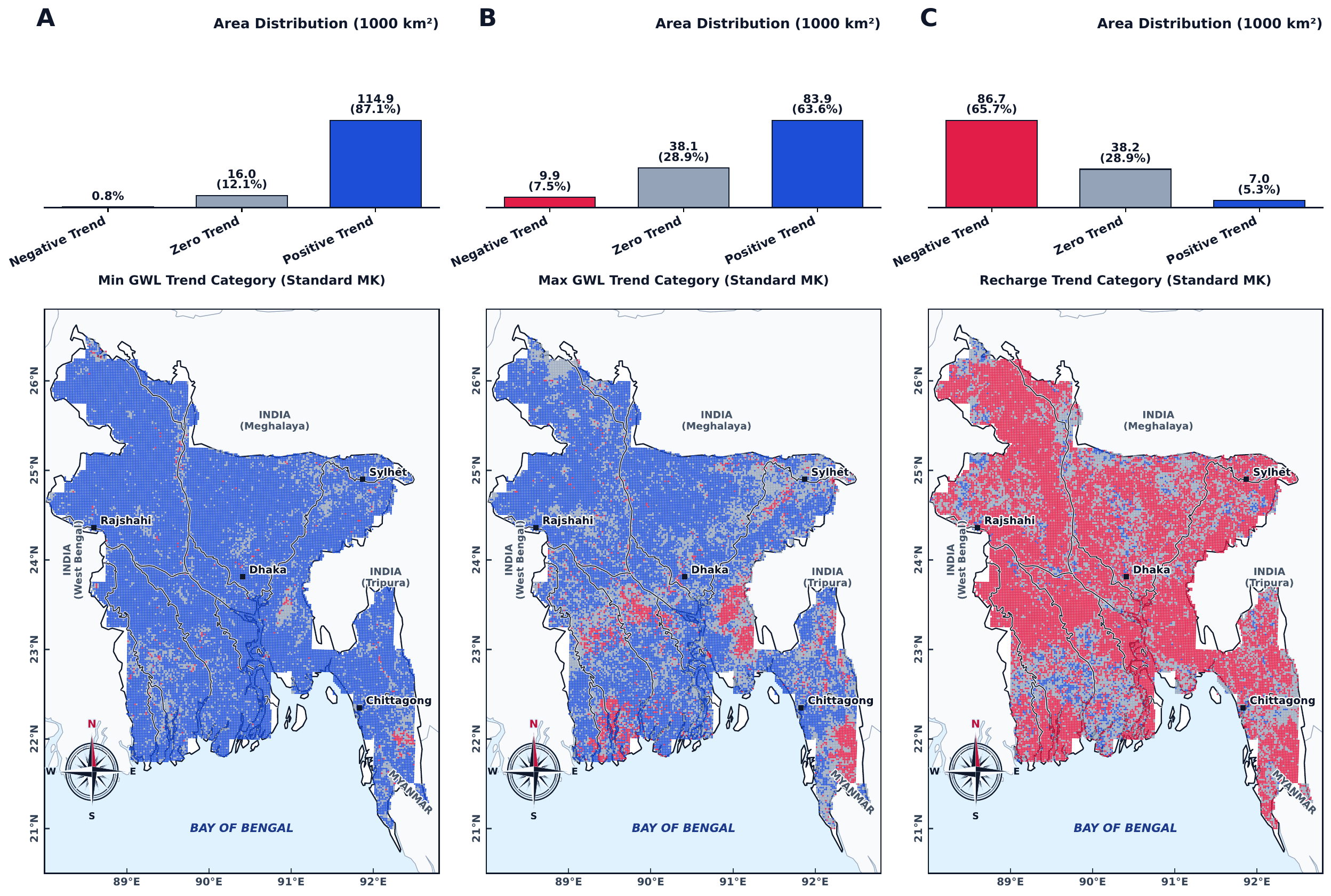}
    \caption{\textbf{Spatial trend significance using standard Mann--Kendall testing:} Conventional Mann--Kendall trend significance across Bangladesh for \textbf{a.} minimum groundwater depth, \textbf{b.} maximum groundwater depth, and \textbf{c.} groundwater recharge. Categories indicate statistically significant negative trends (red), significant positive trends (blue), and non-significant trends (grey).}
    \label{fig:supp_standard_mk}
\end{figure}
\section{Supplementary Methods}

\subsection{Geospatial preprocessing and feature extraction}
\label{sec:geospatial_preprocessing}

Detailed definitions and source datasets for the 15 static hydrogeological features, annual NDVI and NDWI, and groundwater-storage variables are provided in Table~\ref{table:features}. The environmental predictors were prepared at two spatial scales: the approximately \(25~\text{km}\) GLDAS grid used to represent regional conditions and the \(2~\text{km}\) grid used for final groundwater-depth reconstruction. All spatial processing was performed using ArcGIS and Google Earth Engine (GEE) \cite{google_earth_engine}.

Lithology was treated as a categorical feature. At the \(25~\text{km}\) GLDAS scale, the majority lithology within each grid cell was used to represent the regional lithological setting. At the \(2~\text{km}\) scale, the lithology of each target grid cell was retained. Lithology categories were then converted to binary variables using one-hot encoding before model training.

\begin{enumerate}

\item \textbf{Predictors at the GLDAS scale:}

The GLDAS data are provided on an approximately \(0.25^\circ\) grid, corresponding to about \(25~\text{km}\). Continuous environmental predictors were spatially aggregated within each GLDAS grid cell using \texttt{Zonal Statistics}, and the median value was calculated for each cell. This was applied to the continuous HGFs, NDVI, and NDWI, with NDVI and NDWI processed separately for each year. For lithology, the majority category within each GLDAS cell was used.

\item \textbf{Creation of the \(2~\text{km}\) target grid:}

A regular \(2~\text{km} \times 2~\text{km}\) fishnet grid covering Bangladesh was generated using the ArcGIS \texttt{Create Fishnet} tool. The cell width and height were both set to \(2000~\text{m}\), using the WGS 84 / UTM Zone 46N projected coordinate system.

\item \textbf{Predictors at the \(2~\text{km}\) scale:}

The environmental predictors were aggregated to the \(2~\text{km}\) target grid using \texttt{Zonal Statistics as Table}. For each grid cell, the median value of each continuous raster predictor was calculated and joined to the corresponding grid cell using its unique identifier (\texttt{FID}). Lithology was retained as the categorical value for each \(2~\text{km}\) grid cell. The resulting dataset provided the local predictor values used for spatial downscaling.

\item \textbf{Construction of the Student input dataset:}

Each \(2~\text{km}\) grid cell was assigned to its parent \(25~\text{km}\) GLDAS cell using a unique GLDAS grid identifier. The Student dataset was structured with one row for each \(2~\text{km}\) target cell and year. Each row combines the annual GWS and regional HGFs from the parent \(25~\text{km}\) GLDAS cell with the 17 local HGFs of the target cell. Thus, all \(2~\text{km}\) cells within a given GLDAS cell share the same regional GWS and regional HGF values while retaining their own local HGF values. All variables were matched by spatial unit and year before the model dataset was constructed.

After this linkage, grid identifiers and other geographic identifiers were removed from the predictor matrix to prevent the model from learning fixed geographic labels. The final dataset therefore provides each target location with both regional groundwater-storage and hydrogeological information and local hydrogeological characteristics.

\end{enumerate}

\subsection{Model implementation}
\label{sec:model_implementation}

All Teacher and Student models were implemented using the \texttt{RandomForestRegressor} from the Python \texttt{scikit-learn} library, with 100 decision trees (\texttt{n\_estimators=100}) and \texttt{random\_state=42}. All other model parameters were kept at their \texttt{scikit-learn} defaults.

\subsection{Monte Carlo Error Propagation and Trend Robustness}
\label{sec:supp_monte_carlo}

We conducted a 1,000-iteration Monte Carlo simulation to evaluate the robustness of reconstructed groundwater-recharge trends to uncertainty in groundwater-depth predictions and specific yield ($S_y$). The analysis was performed at each grid location over the 2003--2022 period. In each iteration, we perturbed the reconstructed minimum and maximum groundwater depths and specific yield, recalculated annual recharge, and estimated the linear trend in recharge for each location. Trend robustness was assessed from the distribution of simulated trend slopes.

\paragraph{Groundwater-depth prediction uncertainty.}
Prediction uncertainty was represented using the standard deviations of predictions across the 100 trees in the Random Forest model evaluated under Leave-One-District-Out (LODO) cross-validation. For each held-out observation, the uncertainty scales for minimum and maximum groundwater depth were calculated as

\begin{equation}
\sigma_q^{(i,t)}
=
\sqrt{
\frac{1}{B}
\sum_{b=1}^{B}
\left(
\hat{d}_{q,b}^{(i,t)}
-
\overline{\hat{d}}_{q}^{(i,t)}
\right)^2
},
\qquad q\in\{\mathrm{min},\mathrm{max}\},
\end{equation}

where $B=100$ is the number of trees, $\hat{d}_{q,b}^{(i,t)}$ is the prediction from tree $b$, and $\overline{\hat{d}}_{q}^{(i,t)}$ is the mean prediction across the trees. These uncertainty scales represent Random Forest ensemble spread rather than residual prediction errors directly.

To represent systematic location-specific deviations and year-specific variability, standardized perturbations were generated using a weighted combination of two random components:

\begin{align}
\epsilon_{\mathrm{min}}^{(i,t,k)}
&=
0.85\,\eta_{\mathrm{sys,min}}^{(i,k)}
+
0.15\,\eta_{\mathrm{annual}}^{(i,t,k)},\\
\epsilon_{\mathrm{max}}^{(i,t,k)}
&=
0.85\,\eta_{\mathrm{sys,max}}^{(i,k)}
+
0.15\,\eta_{\mathrm{annual}}^{(i,t,k)},
\end{align}

where $i$ denotes grid location, $t$ denotes year, and $k$ denotes Monte Carlo iteration. The systematic components,
$\eta_{\mathrm{sys,min}}^{(i,k)}$ and
$\eta_{\mathrm{sys,max}}^{(i,k)}$, were sampled independently from a standard normal distribution once per location per iteration and retained across years. The annual component,
$\eta_{\mathrm{annual}}^{(i,t,k)}$, was sampled independently for each location-year record and shared between the minimum- and maximum-depth perturbations. All three random components follow a standard normal distribution, $\mathcal{N}(0,1)$, and the weights of 0.85 and 0.15 were applied directly as linear coefficients.

The perturbed groundwater depths were calculated as

\begin{equation}
\hat{d}_{q}^{(i,t,k)}
=
d_q^{(i,t)}
+
\epsilon_q^{(i,t,k)}\sigma_q^{(i,t)},
\qquad q\in\{\mathrm{min},\mathrm{max}\},
\end{equation}

where $d_q^{(i,t)}$ is the baseline reconstructed groundwater depth and $\sigma_q^{(i,t)}$ is the corresponding ensemble-spread estimate. The same location-specific systematic component was retained throughout each simulated time series, while the annual component varied between years.

\paragraph{Specific-yield uncertainty and recharge calculation.}
Specific yield was perturbed independently across locations and Monte Carlo iterations using a normal multiplicative factor with a coefficient of variation of 10\%. Specifically,

\begin{equation}
\hat{S}_y^{(i,k)}
=
S_y^{(i)} Z^{(i,k)},
\qquad
Z^{(i,k)}\sim\mathcal{N}(1,0.10^2),
\end{equation}

where $S_y^{(i)}$ is the baseline specific yield and $\hat{S}_y^{(i,k)}$ is its perturbed value. The same perturbed specific yield was applied to all years at a given location within each iteration.

Annual recharge was recalculated using the water-table fluctuation method:

\begin{equation}
\hat{R}^{(i,t,k)}
=
\left[
\hat{d}_{\mathrm{max}}^{(i,t,k)}
-
\hat{d}_{\mathrm{min}}^{(i,t,k)}
\right]
\hat{S}_y^{(i,k)}
\times 1000,
\end{equation}

where recharge, $\hat{R}$, is expressed in millimetres when groundwater depths are in metres and specific yield is dimensionless. The difference between maximum and minimum groundwater depth represents the annual water-table fluctuation used to estimate recharge.

\paragraph{Trend estimation and robustness classification.}
For each grid location and Monte Carlo iteration, a linear trend was estimated from the simulated annual recharge series using ordinary least-squares linear regression. The trend slope was expressed in $\mathrm{mm\,yr^{-1}}$. Let $\beta^{(i,k)}$ denote the slope estimated for location $i$ in iteration $k$. The resulting distribution of 1,000 slopes was summarized using its mean and the 2.5th and 97.5th percentiles:

\begin{align}
\overline{\beta}^{(i)}
&=
\frac{1}{K}\sum_{k=1}^{K}\beta^{(i,k)},\\
L^{(i)}
&=
Q_{0.025}\left(\beta^{(i,1)},\ldots,\beta^{(i,K)}\right),\\
U^{(i)}
&=
Q_{0.975}\left(\beta^{(i,1)},\ldots,\beta^{(i,K)}\right),
\end{align}

where $K=1{,}000$, $Q_{0.025}$ and $Q_{0.975}$ denote the 2.5th and 97.5th percentiles, and $L^{(i)}$ and $U^{(i)}$ are the bounds of the 95\% simulation interval.

A trend was classified as robust when the simulation interval excluded zero:

\begin{equation}
\mathrm{Robust}^{(i)}
=
\begin{cases}
\mathrm{True}, & L^{(i)}>0 \text{ or } U^{(i)}<0,\\
\mathrm{False}, & L^{(i)}\leq 0\leq U^{(i)}.
\end{cases}
\end{equation}

Robust positive trends were identified when $L^{(i)}>0$, whereas robust negative trends were identified when $U^{(i)}<0$. Locations whose intervals included zero were classified as having uncertain trend direction. The simulation interval characterizes the distribution of trends under the specified perturbation assumptions and is not a conventional confidence interval from a formal statistical trend test.

\subsection{Modified Mann--Kendall Trend Test}
\label{sec:supp_mmk}

Long-term trends in annual groundwater-recharge time series were assessed using the Modified Mann--Kendall (MMK) test to account for temporal autocorrelation, which can affect the statistical significance of conventional Mann--Kendall trend tests \cite{hamed1998modified,yue2002modified}. The procedure detrends each time series, evaluates serial dependence in the detrended residuals, and adjusts the variance of the Mann--Kendall statistic when significant autocorrelation is detected ($p^*<0.05$).

For a time series $x_1,\ldots,x_n$, the Mann--Kendall statistic is

\begin{equation}
S=\sum_{i=1}^{n-1}\sum_{j=i+1}^{n}
\operatorname{sign}(x_j-x_i),
\end{equation}

where

\begin{equation}
\operatorname{sign}(x_j-x_i)=
\begin{cases}
+1, & x_j-x_i>0,\\
0, & x_j-x_i=0,\\
-1, & x_j-x_i<0.
\end{cases}
\end{equation}

The statistic measures the direction of the monotonic trend. Its variance, $V(S)$, is adjusted for serial dependence when the autocorrelation assessment indicates that a correction is required. The resulting standardized test statistic is used to determine the two-sided significance of the trend. The trend direction is determined by the sign of $S$, while statistical significance is assessed using the corrected test statistic and its associated $p$-value.

The MMK test was used to identify statistically significant long-term recharge trends after accounting for temporal autocorrelation. This analysis is distinct from the Monte Carlo robustness assessment, which evaluates the sensitivity of local linear trend slopes to uncertainty in reconstructed groundwater depths and specific yield.

\begin{figure}[!htbp]
    \centering
     \includegraphics[width=\textwidth]{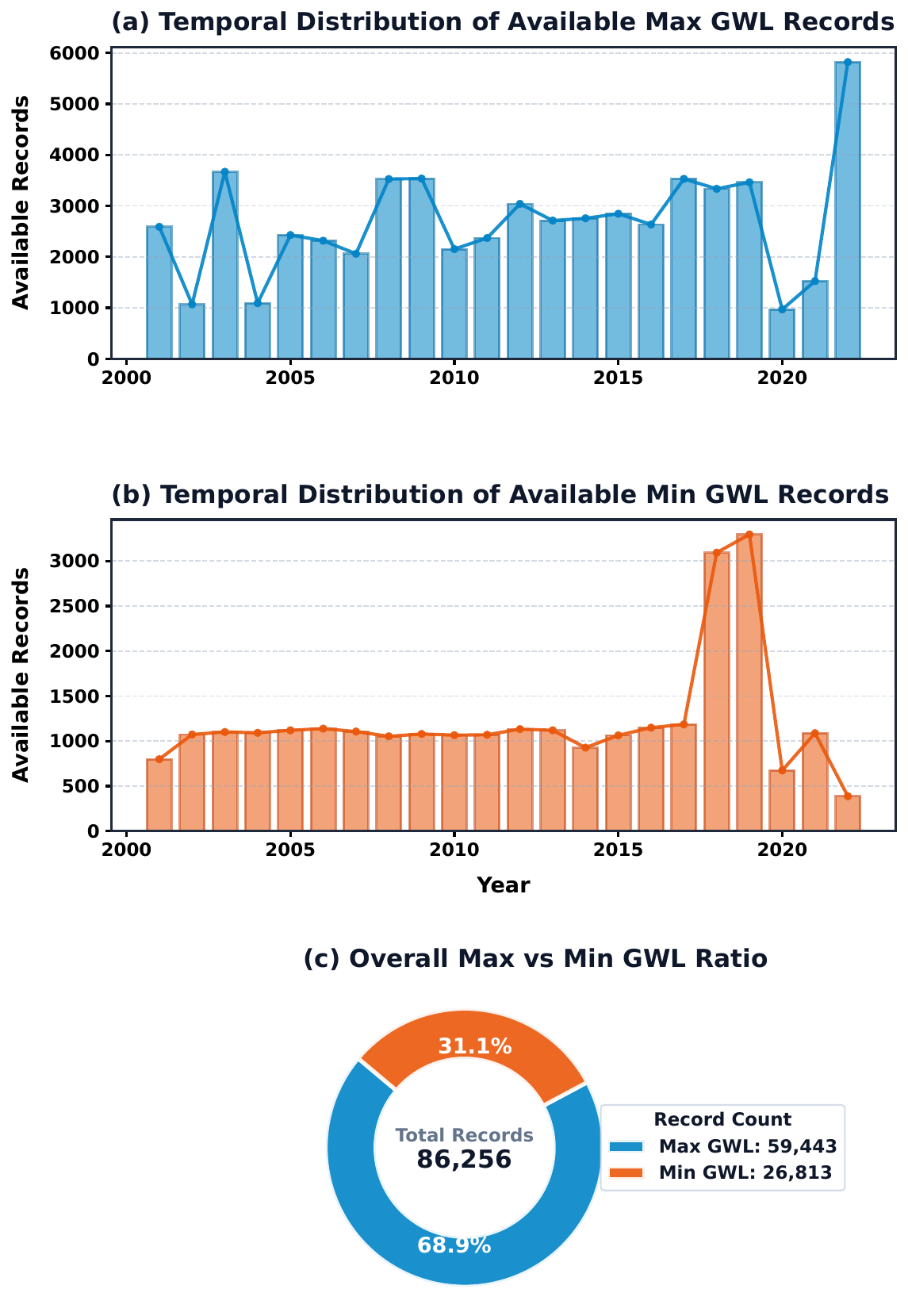}
    \caption{\textbf{Data Distribution:} \textbf{A, B.} Annual maximum and minimum groundwater levels. \textbf{C.} Combined multi-year data showing maximum depths exceed minimums (below ground level).}
    \label{fig:distribution}
\end{figure}

\section{SHAP Value Computation for Tree-Based Models}
\label{subsec:tree_shap}

We used SHAP (SHapley Additive exPlanations) values~\cite{van2022tractability} to understand how the environmental predictors influenced the predictions of the tree-based models, including Random Forest, Gradient Boosted Trees and XGBoost. SHAP provides a way to quantify how much each predictor contributes to an individual model prediction relative to the model's average prediction.

SHAP values were calculated using the \texttt{TreeExplainer} method~\cite{shap_treeexplainer}, which is specifically designed for tree-based models. Instead of repeatedly refitting the model with different combinations of predictors, TreeExplainer uses the structure of the fitted trees to calculate the contributions efficiently.

\subsubsection*{Mathematical Formulation}

For an input observation

$$
\mathbf{x}=[x_1,x_2,\ldots,x_M],
$$

where $M$ is the number of predictors, let $f(\mathbf{x})$ represent the model prediction. SHAP expresses this prediction as

\begin{equation}
f(\mathbf{x})=\phi_0+\sum_{i=1}^{M}\phi_i,
\label{eq:tree_shap_decomposition}
\end{equation}

where $\phi_0$ is the model's average prediction and $\phi_i$ is the contribution of predictor $i$ to the prediction.

Thus,

\begin{equation}
f(\mathbf{x})-\phi_0=\sum_{i=1}^{M}\phi_i.
\label{eq:tree_shap_contribution}
\end{equation}

In simple terms, the difference between an individual prediction and the model's average prediction can be divided among the input predictors. The SHAP value of each predictor indicates the direction and magnitude of its contribution.

The general SHAP formulation considers the contribution of a predictor across different combinations of the other predictors:

\begin{equation}
\phi_i =
\sum_{S\subseteq F\setminus{i}}
\frac{|S|!(M-|S|-1)!}{M!}
\left[
v(S\cup{i})-v(S)
\right],
\label{eq:shapley_general}
\end{equation}

where $F$ is the set of all predictors, $S$ represents a subset of predictors, and $v(S)$ represents the model output associated with that subset. TreeExplainer calculates these contributions efficiently by using the structure of the decision trees rather than explicitly evaluating every possible combination.

\subsubsection*{Interpretation of SHAP Values}

The sign of a SHAP value indicates the direction of a predictor's influence on a particular prediction. A positive SHAP value ($\phi_i>0$) means that the predictor pushes the model prediction above the average prediction, whereas a negative SHAP value ($\phi_i<0$) means that it pushes the prediction below the average. Larger absolute values indicate a stronger influence.

We examined SHAP values across all modelled observations to assess the overall importance of the predictors. Summary plots were used to show the distribution and magnitude of predictor contributions across observations. Dependence plots were used to examine how the contribution of a predictor changed as its value increased or decreased.

These analyses allowed us to identify which environmental predictors had the greatest influence on the groundwater-level predictions and to examine the direction of their influence, providing an interpretable view of how the machine-learning models used the available environmental information.

\begin{figure}
    \centering
    \includegraphics[width=\linewidth]{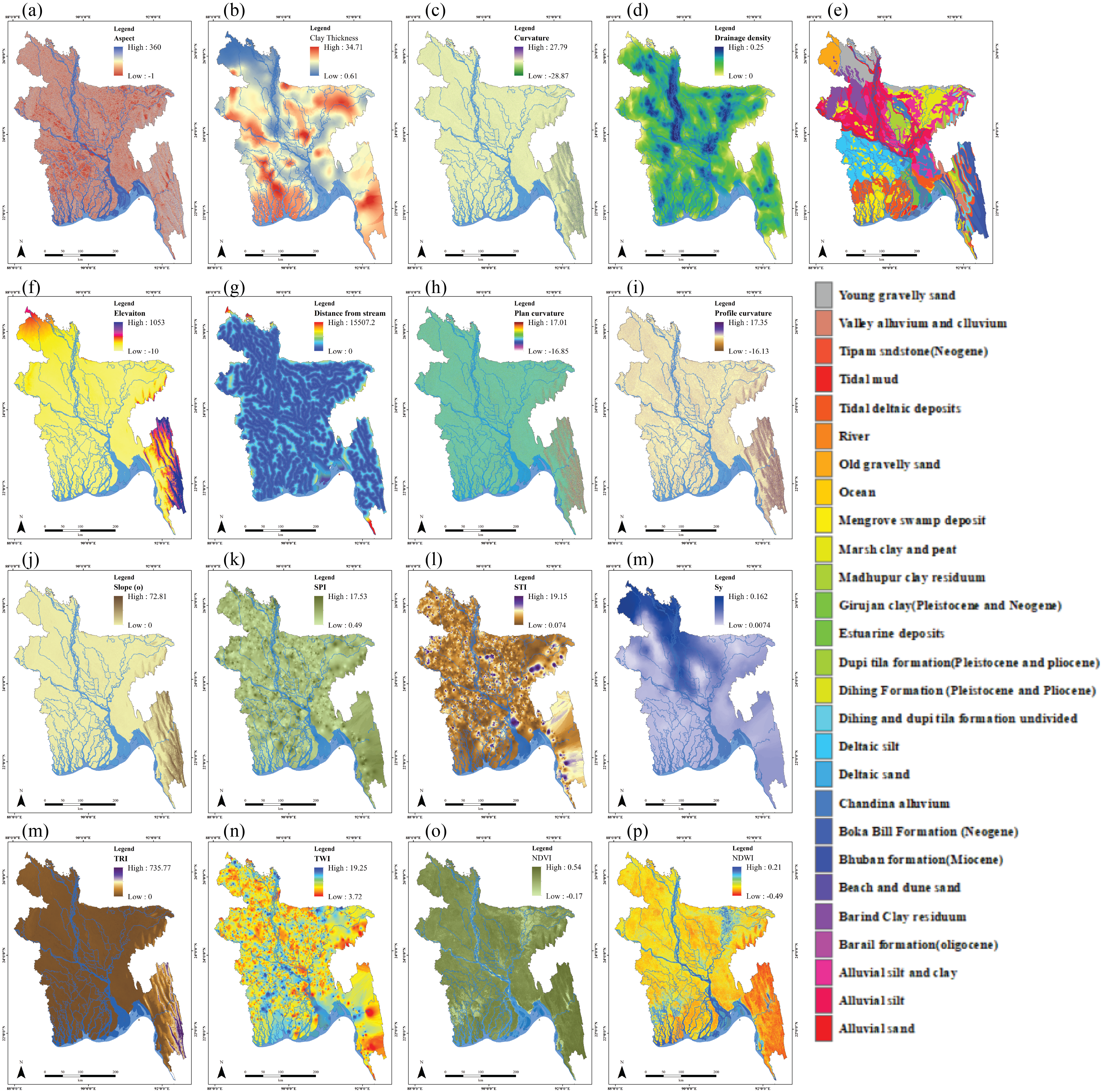}
   \caption{\textbf{Hydro-Geological Feature Map:} Distribution of 17 hydro-geological feature (HGF) values across the country.}
    \label{fig:hgf_dist}
\end{figure}

\begin{longtable}{|p{3cm}|p{3cm}|p{5cm}|p{3cm}|}
    \caption{\textbf{Summary of Numerical and Categorical Factors:} First 17 features are referred to as Hydro-Geological Features (HGFs) in the text.} \label{tab:numerical_categorical_factors} \\
    \hline
    \textbf{Features (data type)} & \textbf{Characteristics} & \textbf{Equation/ Tool name} & \textbf{Data Source} \\
    \hline
    \endfirsthead
    
    % \hline
    % \textbf{Features (data type)} & \textbf{Characteristics} & \textbf{Equation/ Tool name} & \textbf{Data Source} \\
    % \hline
    % \endhead

    \hline
    \endfoot
    Slope (Numerical) & The steepness of the terrain, derived from the DEM. Slope influences surface runoff, soil erosion, and infiltration \cite{gao2024comprehensive}.& \[\text{Slope\%}
     = 
    100 \times \frac{\sqrt{A X^2 + B Y^2}}{\text{Pixel Size (DEM)}}
    \]
    
    \noindent
    \textbf{Where:}
    \begin{itemize}
        \item \(A X\) = Filtered DEM with x-gradient filter
        \item \(B Y\) = Filtered DEM with y-gradient filter
        \item \text{Pixel Size (DEM)} = Pixel size of the DEM
    \end{itemize}
    \cite{ritter2007species}
    \break
    \textbf{Note:} Derived from slope of 
    Spatial Analyst Tool in 
    ArcGIS Pro platform
    &
    Derived from Shuttle 
    Radar Topography 
    Mission (SRTM) \cite{srtm_nasa} 
    based digital 
    elevation model 
    (SRTM-DEM) 
    of 30 m resolution\\
    \hline
    Drainage Density (Numerical) & The total length of streams per unit area of the watershed. Higher drainage density indicates better drainage and less water retention in the soil \cite{anderson2010geomorphology}. &\[
    D_d = \frac{\sum_{i=1}^{n} D_i}{A}
    \]
    
    \noindent
    \textbf{Where:}
    \begin{itemize}
        \setlength\itemsep{-0.3em} % Reduce spacing between items
        \item \(D_i\): Total length of streams
        \item \(A\): Area of a grid
        \item \(D_d\): Drainage Density
    \end{itemize}
    \cite{horton1932drainage}
    \break
    \textbf{Note:} Derived from flow 
    accumulation, stream 
    order, fill, etc. of Spatial 
    Analyst Tool in ArcGIS 
    Pro platform 
    &
    Derived from 
    SRTM-DEM of 30 m 
    resolution \\
    \hline
    Elevation (Numerical) & The height above sea level, derived from the DEM. Elevation influences climatic and hydrological conditions \cite{kenward2000effects}.&& Derived from 
    SRTM-DEM of 30 m 
    resolution  \\
    \hline
    Distance from Stream (Numerical) & The proximity of a location to the nearest stream, which affects the availability of water and the potential for flooding \cite{suwanno2023gis}.&\[
    d_{ij} = \sqrt{\sum_{k=1}^{n} (X_{ik} - Y_{jk})^2}
    \]
    
    \noindent
    \textbf{Where:}
    \begin{itemize}
        \setlength\itemsep{-0.3em} % Reduce spacing between items
        \item \(d_{ij}\): Distance from stream for \(i, j\) locations
        \item \(k\): Represents features
    \end{itemize}
    
    \noindent
    \textbf{Note:} Used Euclidean distance of Spatial Analyst Tool to calculate in ArcGIS Pro platform.
     &
    Derived from 
    SRTM-DEM of 30 m 
    resolution\\
    \hline
    Topographic Wetness Index (TWI) (Numerical) & A measure of topographic control on hydrological processes, calculated using slope and upstream contributing area. Higher values indicate wetter areas.& \[
    TWI = \ln\left(\frac{A_s}{\beta}\right)
    \]
    
    \noindent
    \textbf{Where:}
    \begin{itemize}
        \setlength\itemsep{-0.3em} % Reduce spacing between items
        \item \(A_s\): Catchment area per unit
        \item \(\beta\): Slope angle in degrees
        \item \(TWI\): Topographic Wetness Index
    \end{itemize}
    
    \noindent
    \textbf{Note:} Calculated from flow accumulation and slope using the Spatial Analyst Tool in ArcGIS Pro platform.
     &\\
    \hline
    Terrain Ruggedness Index (TRI) (Numerical) & Quantifies the terrain's ruggedness by measuring elevation differences between adjacent grid cells. Rugged terrain affects runoff and groundwater recharge \cite{grinevskii2014effect,verma2024spatial}.&\[
    R = \sqrt{\text{max}^2 - \text{min}^2}
    \]
    
    \noindent
    \textbf{Where:}
    \begin{itemize}
        \setlength\itemsep{-0.3em} % Reduce spacing between items
        \item \(\text{max}\): The highest cell value in the DEM
        \item \(\text{min}\): The lowest cell value in the DEM
    \end{itemize}
    
    \noindent
    \textbf{Note:} Calculated using ArcGIS Pro platform.&
    Derived from 
    SRTM-DEM of 30 m 
    resolution
     \\
    \hline
    Stream Topographic Index (STI) (Numerical) & Indicates the likelihood of a location being affected by stream flow based on elevation and proximity to streams \cite{infascelli2013testing}.&\[
    STI = \left(\frac{A_s}{22.13}\right)^{0.6} \times \left(\frac{\sin \beta}{0.0896}\right)^{1.3}
    \]
    
    \noindent
    \textbf{Where:}
    \begin{itemize}
        \setlength\itemsep{-0.3em} % Reduce spacing between items
        \item \(A_s\): Catchment area per unit
        \item \(\beta\): Slope angle in degrees
        \item \(STI\): Stream Topographic Index
    \end{itemize}
    
    \noindent
    \textbf{Note:} Calculated using ArcGIS Pro platform.&
    Derived from 
    SRTM-DEM of 30 m 
    resolution
     \\
    \hline
    Stream Power Index (SPI) (Numerical) & Represents the erosive power of flowing water, calculated using flow accumulation and slope. High SPI values suggest areas with significant stream energy \cite{leelaruban2017examining}.&\[
    SPI = A_s \times \tan \beta
    \]
    
    \noindent
    \textbf{Where:}
    \begin{itemize}
        \setlength\itemsep{-0.3em} % Reduce spacing between items
        \item \(A_s\): Catchment area per unit
        \item \(\beta\): Slope angle in degrees
        \item \(SPI\): Stream Power Index
    \end{itemize}
    
    \noindent
    \textbf{Note:} Calculated using ArcGIS Pro platform.&
    Derived from 
    SRTM-DEM of 30 m 
    resolution
     \\
    \hline
    Profile Curvature,  Plan Curvature \& Curvature (Numerical) & Measures the curvature of the land surface in the direction of the slope. It affects the acceleration or deceleration of water flow \cite{arcgis_curvature}. influencing water flow and sediment transport \cite{peckham2011profile}&\[
    \kappa = \left|\frac{d\mathbf{T}}{ds}\right|
    \]
    
    \noindent
    \textbf{Where:}
    \begin{itemize}
        \setlength\itemsep{-0.3em} % Reduce spacing between items
        \item \(\mathbf{T}\): Unit tangent vector
        \item \(ds\): Differential of the curve length
        \item \(|\cdot|\): Represents the magnitude of the vector
    \end{itemize}
    
    \noindent
    \textbf{Note:} Calculated using the Curvature tool of the Spatial Analyst extension in ArcGIS Pro software.&
    Derived from 
    SRTM-DEM of 30 m 
    resolution
     \\
    \hline
    Aspect (Numerical) & The compass direction of the slope, affecting sun exposure, evaporation, and vegetation patterns.&Aspect tool in ArcGIS 
    Pro Software.  &
    Derived from 
    SRTM-DEM of 30 m 
    resolution\\
    \hline
    Specific Yield (Sy) (Numerical) & The ratio of water that can be drained by gravity to the total volume of the aquifer. It is a critical factor for estimating groundwater availability \cite{lv2021comprehensive}.&Ordinary kriging of 
Spatial Analyst tool in 
ArcGIS Pro Software &\cite{shamsudduha2022bengal} \\
    \hline
    Lithology Clay Thickness  (Numerical) & Represents the thickness of clay layers in the lithology, which impacts water retention and aquifer properties \cite{xing2018blocking}.&&\cite{shamsudduha2011impact} \\
    \hline
    Lithology (Categorical) & Represents the geological characteristics of soil and rock types in the study area. Lithology influences permeability, porosity, and groundwater flow. This factor is one-hot encoded to transform categorical lithological data into a numerical format suitable for analysis.&Digitizing in ArcGIS Pro software. This factor is one-hot encoded to transform categorical lithological data into a numerical format suitable for 
    analysis.&\cite{geological1990geological}   \\
    \hline
    Yearly mean Normalized Difference Vegetation Index (NDVI)
    (Numerical)& Represents the average level 
    of vegetation greenness for 
    a given year, serving as a 
    key indicator of plant 
    health, density, and 
    productivity. & \[
    \text{NDVI} = \frac{\text{NIR} - \text{Red}}{\text{NIR} + \text{Red}}
    \]

    \textbf{Where:}
    \begin{itemize}
        \item \(NIR\) = Near Infrared \end{itemize}
    
    \noindent
    \textbf{Note:} Calculated using the Google Earth Engine (GEE) \cite{google_earth_engine} platform.
    &Landsat 5 \cite{landsat5} and Landsat 8 \cite{landsat8}
    \\
    \hline
    Yearly mean Normalized Difference Water Index (NDWI)
    (Numerical)&Represents the average level 
    of surface water presence 
    for a given year, serving as 
    a key indicator of water 
    bodies and extent. & \[
    \text{NDWI} = \frac{\text{NIR} - \text{SWIR}}{\text{NIR} + \text{SWIR}}
    \]
\textbf{Where:}
    \begin{itemize}
    \item \(NIR\) = Near Infrared
     \item \(SWIR\) = Short Wave Infrared
    \end{itemize}
   
    \noindent
    \textbf{Note:} Calculated using the Google Earth Engine (GEE) platform.
    &Landsat 5 \cite{landsat5} and Landsat 8 \cite{landsat8}
    \\
    \hline
    groundwater storage (Numerical)&Represents the average level 
    of groundwater storage 
    for a given year, which represent the total volume of water stored in aquifers within a 25 km × 25 km grid. groundwater storage serves as an indirect indicator for groundwater depth: increased storage correlates with shallower water tables (shallower depth below ground level), while decreased storage implies deeper water tables (greater depth below ground level). &Downloaded and 
    Calculated in 
    Google Earth 
    Engine (GEE) 
    platform  & Global Land Data 
    Assimilation 
    Systems (GLDAS) at 
    0.25 x 0.25 degree 
    \label{table:features}
\end{longtable}

%=============================================%%
% For submissions to Nature Portfolio Journals %%
% please use the heading ``Extended Data''.   %%
%=============================================%%

%=============================================================%%
% Sample for another appendix section			       %%
%=============================================================%%

% \section{Example of another appendix section}\label{secA2}%
% Appendices may be used for helpful, supporting or essential material that would otherwise 
% clutter, break up or be distracting to the text. Appendices can consist of sections, figures, 
% tables and equations etc.

%===========================================================================================%%
% If you are submitting to one of the Nature Portfolio journals, using the eJP submission   %%
% system, please include the references within the manuscript file itself. You may do this  %%
% by copying the reference list from your .bbl file, paste it into the main manuscript .tex %%
% file, and delete the associated \verb+\bibliography+ commands.                            %%
%===========================================================================================%%


\end{document}